\documentclass[lettersize,journal,compsoc]{IEEEtran}
\usepackage{amsmath,amsfonts}
\usepackage[ruled,vlined,linesnumbered]{algorithm2e}
\SetKwInput{KwIn}{Input}
\SetKwInput{KwOut}{Output}
\usepackage{array}
\usepackage[caption=false,font=normalsize,labelfont=sf,textfont=sf]{subfig}
\usepackage{textcomp}
\usepackage{stfloats}
\usepackage{url}
\usepackage{verbatim}
\usepackage{graphicx}
\usepackage{cite}
\usepackage{multirow}
\usepackage{pifont}
\usepackage{microtype}
\usepackage{booktabs}
\usepackage[table]{xcolor}
\usepackage[format=plain,labelformat=simple,labelsep=period,font=small,compatibility=false]{caption}
\usepackage[font=footnotesize,skip=3pt,subrefformat=parens]{subcaption}
\usepackage[pagebackref=true,breaklinks=true, colorlinks,bookmarks=false,linkcolor=blue,citecolor=blue]{hyperref}

\newcommand{\best}[1]{\cellcolor[HTML]{F8B5B4}{#1}}
\newcommand{\second}[1]{\cellcolor[HTML]{FAD8B5}{#1}}
\newcommand{\third}[1]{\cellcolor[HTML]{FDFFB6}{#1}}

\def\eg{\emph{e.g.}}

\begin{document}

\title{OmniVTON++: Training-Free Universal Virtual Try-On with Principal Pose Guidance}

\author{
Zhaotong Yang,
Yong Du,
Shengfeng He,
Yuhui Li,
Xinzhe Li,
Yangyang Xu,
Junyu Dong,
and Jian Yang

\IEEEcompsocitemizethanks{
                
\IEEEcompsocthanksitem Corresponding author: Yong Du.

		\IEEEcompsocthanksitem Zhaotong Yang is with the School of Computer Science and Technology, Ocean University of China, Qingdao, China, and the School of Computer Science and Engineering, Nanjing University of Science and Technology, Nanjing, China. E-mail: ztyang808@njust.edu.cn.

        \IEEEcompsocthanksitem Yong Du is with the School of Computer Science and Technology, Ocean University of China, Qingdao, China, and the Sanya Oceanographic Institution, Ocean University of China, Sanya, China. E-mail: csyongdu@ouc.edu.cn. 
        
        \IEEEcompsocthanksitem Shengfeng He is with the School of Computing and Information Systems, Singapore Management University, Singapore. E-mail: shengfenghe@smu.edu.sg.
        
        \IEEEcompsocthanksitem Yuhui Li, Xinzhe Li, and Junyu Dong are with the School of Computer Science and Technology, Ocean University of China, Qingdao, China. E-mail: liyuhui1150@stu.ouc.edu.cn, lixinzhe@stu.ouc.edu.cn, dongjunyu@ouc.edu.cn.        		
		
		\IEEEcompsocthanksitem	Yangyang Xu is with the School of Intelligence Science and Engineering, Harbin Institute of Technology, Shenzhen, China. E-mail: xuyangyang@hit.edu.cn.
		
		\IEEEcompsocthanksitem	Jian Yang is with the School of Computer Science and Engineering, Nanjing University of Science and Technology, Nanjing, China. E-mail: csjyang@njust.edu.cn.}
}

\markboth{}
{Yang \MakeLowercase{\textit{et al.}}: OmniVTON++: Training-Free Universal Virtual Try-On with Principal Pose Guidance}

\IEEEcompsoctitleabstractindextext{
\begin{abstract}
Image-based Virtual Try-On (VTON) concerns the synthesis of realistic person imagery through garment re-rendering under human pose and body constraints. In practice, however, existing approaches are typically optimized for specific data conditions, making their deployment reliant on retraining and limiting their generalization as a unified solution. We present OmniVTON++, a training-free VTON framework designed for universal applicability. It addresses the intertwined challenges of garment alignment, human structural coherence, and boundary continuity by coordinating Structured Garment Morphing for correspondence-driven garment adaptation, Principal Pose Guidance for step-wise structural regulation during diffusion sampling, and Continuous Boundary Stitching for boundary-aware refinement, forming a cohesive pipeline without task-specific retraining. Experimental results demonstrate that OmniVTON++ achieves state-of-the-art performance across diverse generalization settings, including cross-dataset and cross-garment-type evaluations, while 
reliably operating across scenarios and diffusion backbones within a single formulation. In addition to single-garment, single-human cases, the framework supports multi-garment, multi-human, and anime character virtual try-on, expanding the scope of virtual try-on applications. The code is available at \url{https://github.com/Jerome-Young/OmniVTON-PlusPlus}.
\end{abstract}

}

\maketitle

\IEEEdisplaynotcompsoctitleabstractindextext
\IEEEpeerreviewmaketitle

\section{Introduction}
\label{sec:intro}
\IEEEPARstart{I}{mage}-virtual try-on (VTON) digitally dresses a person with a given garment, aiming to generate visually realistic results that preserve garment appearance while remaining consistent with human structure. Achieving this requires resolving complex interactions among garment geometry, human pose, and visual continuity, where minor misalignments readily produce perceptible artifacts. Such sensitivities influence users' judgment and decision confidence in online apparel visualization, making VTON a technically demanding yet impactful tool for reducing uncertainty and lowering return rates in e-commerce.

\begin{figure*}
    \centering
    \includegraphics[width=\textwidth]{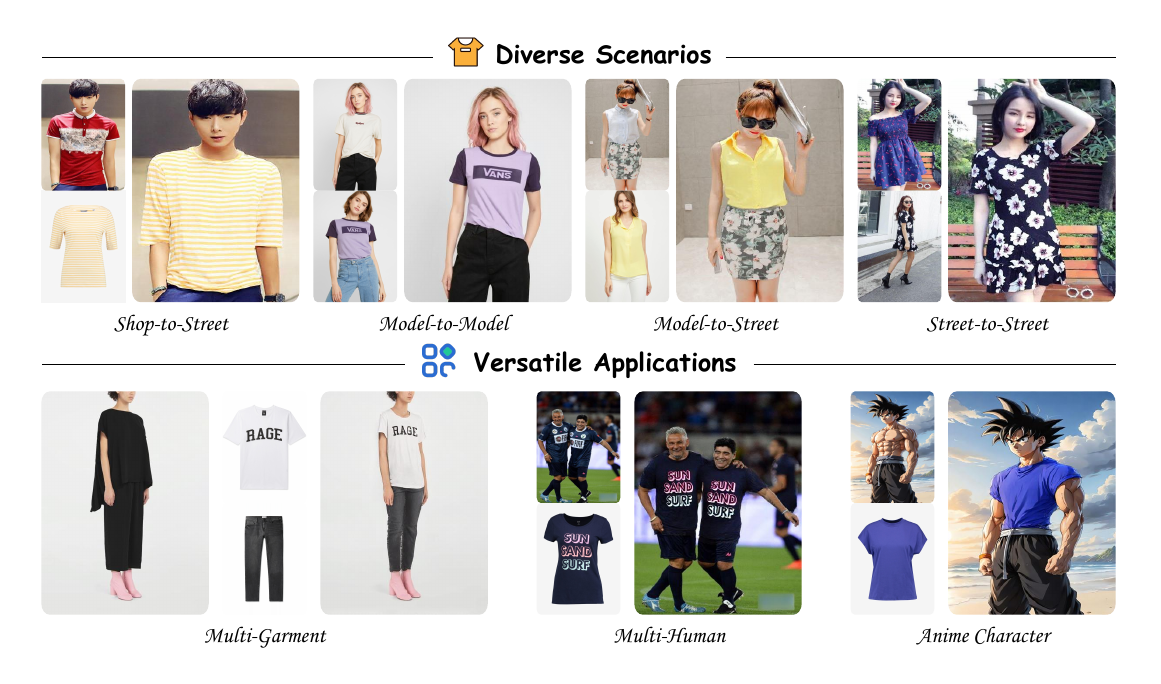}
    \captionof{figure}{We present OmniVTON++, a training-free framework that unifies virtual try-on from in-shop to in-the-wild. It accommodates varied garment sources and person inputs, maintaining consistent garment geometry, texture, and human structure. Notably, OmniVTON++ unlocks multi-garment, multi-human, and anime virtual try-on beyond traditional real-person single-instance settings, all without additional tuning.}
    \label{fig:teaser} 
\end{figure*}

Most existing VTON methods are developed under narrowly specified training regimes, where model behavior is constrained by garment domains and person image acquisition conditions. The field has converged on two paradigms: in-shop and in-the-wild settings. In-shop approaches~\cite{hr-vton,vitonhd,dci-vton,ootd} assume controlled indoor model imagery with simplified backgrounds and lighting, and typically rely on paired supervision under the Shop-to-Model setting. In contrast, in-the-wild methods~\cite{street-tryon} operate on person images captured in unconstrained scenes, with varying background complexity and illumination, such as in Shop-to-Street and Model-to-Street scenarios. Nevertheless, both paradigms exhibit limited robustness to distribution shifts and therefore require retraining for adaptation, restricting their applicability under unseen data conditions.

Breaking away from training-dependent design necessitates rethinking the formulation of virtual try-on. In a training-free setting, garment transformation and human synthesis can no longer draw on priors encoded in task-specific learning, but instead must be realized through mechanisms operating within the generation process. This departure exposes fundamental challenges in such a setting.

One such challenge lies in preserving fine-grained garment attributes while enforcing geometric consistency. In the absence of learned garment–body correspondence, the deformation patterns underpinning appearance-consistent alignment are not explicitly defined, making delicate garment characteristics difficult to retain during geometric transformation.

A second challenge concerns ensuring coherent human structure in the generated results. Pose representations such as keypoints~\cite{openpose} or DensePose maps~\cite{densepose} are widely used in virtual try-on, yet their integration often depends on retraining to couple pose cues with image synthesis. When such adaptation is unavailable, structural inconsistencies become prevalent, particularly for garments that impose limited constraints on body shape, such as sleeveless vests.

To tackle these challenges, we propose OmniVTON++, a virtual try-on framework that casts garment adaptation and human structure control into a training-free, diffusion-based formulation using off-the-shelf models. To handle garment deformation under geometric alignment, we introduce Structured Garment Morphing (SGM). SGM assumes the availability of a target-garment-wearing person image that supplies parseable human structure for correspondence modeling. When the garment is provided in an in-shop form, and such an image is not available, SGM synthesizes a garment-contained pseudo-person image to fulfill this role. It then establishes multi-part semantic correspondences between the garment-contained image and the given person image using human parsing and skeletal priors. Based on the inferred correspondences, SGM applies region-wise garment morphing through part-specific transformations, yielding structurally coherent garment adaptation while preserving fine-grained appearance.

We also introduce Principal Pose Guidance (PPG), a step-wise pose guidance mechanism defined over diffusion sampling for human structure control. Naively injecting pose information, for example through DDIM inversion~\cite{ddim}, preserves structural cues yet often transfers residual garment appearance from the source clothing. PPG instead constructs a proxy latent that encodes pose-consistent structure while excluding original clothing appearance. During diffusion sampling, PPG modulates the noise at each step, computing residuals relative to the proxy latent along the principal components of the intermediate prediction. By operating on a designated subspace, pose consistency is enforced while retaining sufficient degrees of freedom for appearance-related conditions to guide garment synthesis.

As a consequence of part-wise processing in SGM, visual discontinuities may emerge along garment boundaries. We therefore introduce Continuous Boundary Stitching (CBS), a complementary refinement mechanism applied during generation. CBS captures feature interactions between the latent representations of the garment image and the garment-infused image, encouraging smooth transitions across adjacent garment regions and promoting visual coherence in the synthesized results.

We conduct extensive experiments to assess OmniVTON++ under cross-dataset and cross-garment-type generalization, as well as multiple virtual try-on scenarios and diffusion backbones. Across these evaluations, OmniVTON++ maintains stable results under variations in data sources, garment categories, application scenarios, and model architectures without retraining. Beyond standard single-garment, single-human virtual try-on, the framework is further examined on multi-garment, multi-human, and anime character settings, covering complex try-on configurations within the same training-free formulation.

The main contributions of this work are summarized as follows:
\begin{itemize}
\item We propose OmniVTON++, a training-free diffusion-based framework that enables universal virtual try-on across both in-shop and in-the-wild scenarios.
\item We introduce Structured Garment Morphing, which establishes garment–body correspondence for anatomically consistent garment adaptation while preserving fine-grained appearance.
\item We develop Principal Pose Guidance, a step-wise pose control mechanism acting during diffusion sampling for preserving coherent human structure, together with Continuous Boundary Stitching to improve visual continuity along garment boundaries.
\item We achieve state-of-the-art performance under cross-dataset and cross-garment-type generalization, and maintain stable results across diverse virtual try-on scenarios and diffusion backbones. Our framework further supports multi-garment, multi-human, and anime character virtual try-on.
\end{itemize}

A preliminary version of this work was published at ICCV 2025 as OmniVTON~\cite{omnivton}. The present journal version, OmniVTON++, introduces several major extensions. First, we introduce PPG to provide step-wise pose control throughout diffusion sampling, moving beyond the initialization-level pose guidance used in OmniVTON and enabling more precise human structural regulation. Second, within SGM, we adopt a garment-centric virtual dressing step for in-shop garments, replacing the original attention-modulated outpainting for correspondence estimation and yielding more reliable structural cues when human context is absent. Third, we extend CBS to diffusion transformers as CBS-DiT by introducing a positional index realignment mechanism that resolves RoPE-induced positional ambiguity under multi-image conditioning, eliminating erroneous attention alignments that cause visible artifacts. Finally, we expand the experimental study, including renewed quantitative and qualitative evaluations, robustness analysis across diffusion backbones, and broader ablation analyses under alternative designs, as well as demonstrations of applicability outside standard settings, including multi-garment and anime character virtual try-on.

\section{Related work}
\label{sec:rela}

\subsection{Image-Based Virtual Try-On}
Existing image-based virtual try-on methods~\cite{cat-dm,stableviton,d4-vton,ladi-vton,ootd,wear-any-way} are commonly organized by the person image domain, which largely dictates scene complexity. Under this view, the literature is typically divided into in-shop scenarios, where garments are transferred onto person images captured in controlled indoor environments, and in-the-wild scenarios, where person images are drawn from unconstrained real-world scenes with cluttered backgrounds and illumination changes.

Early works~\cite{viton,vitonhd,hr-vton,clothflow,styleflow,dci-vton} predominantly focused on the Shop-to-Model setting, which represents the most standard in-shop configuration, transferring flat-lay garment images onto indoor model images under paired supervision. IDM-VTON~\cite{idm-vton} and StableVITON~\cite{stableviton} extend in-shop pipelines to in-the-wild scenarios using attention-based designs, and their alignment quality decreases when the garment input changes from flat-lay images to person-worn forms. PastaGAN~\cite{pastagan} and PastaGAN++~\cite{pastagan++} support Model-to-Model virtual try-on through patch-routed disentanglement, yet they are restricted to person-worn garment inputs and do not cover flat-lay garments or in-the-wild scenes. More recent efforts, such as StreetTryOn~\cite{street-tryon}, attempt to accommodate both in-shop and in-the-wild scenarios within a unified framework; their performance remains dependent on the training protocol, exhibiting limited transfer across domains and garment categories.

Overall, existing methods rely on training-coupled supervision, which often confines a model to a narrow operating regime and reduces adaptability under shifts in input conditions. In contrast, OmniVTON++ formulates virtual try-on in a training-free manner, allowing a single framework to operate across varied garment representations and person image domains.

\subsection{Garment Warping}
Garment warping aims to deform a garment image to align with the geometry of a given person prior to appearance synthesis. Existing methods can be broadly divided into implicit and explicit paradigms. Implicit methods~\cite{cat-dm,stableviton,idm-vton,ladi-vton,ootd,wear-any-way} leverage diffusion models to jointly model garment deformation and human synthesis within a unified generative process. Despite their flexibility, the absence of explicit geometric constraints often leads to garment misalignment and texture inconsistency, especially in unconstrained scenes.

In contrast, explicit methods model garment geometry directly, offering more controlled alignment and structural consistency. Thin Plate Spline (TPS)~\cite{tps} based transformations~\cite{viton,vitonhd} use sparse control points and are insufficient under large pose variations. Flow based formulations~\cite{clothflow,styleflow,gp-vton,d4-vton} enable pixel level alignment through dense optical flow estimation~\cite{flow}, yielding improved correspondence quality. Nevertheless, these formulations depend on paired training data and are tuned to particular training protocols, limiting their scalability across varied garment representations and application settings.

While out-of-distribution generalization~\cite{song2022editing} and domain-agnostic modeling~\cite{du2025one} have attracted increasing interest in other vision tasks, such considerations remain less developed for garment warping. OmniVTON++ advances this direction by enabling training-free, universal garment adaptation via explicit correspondence-driven deformation.

\subsection{Exemplar-Based Image Inpainting}
Exemplar-based image inpainting~\cite{pbe,tigic,anydoor,cross-image,storydiffusion,freecustom,textual-inversion,e4t} studies reference-guided appearance completion for missing regions, encompassing both generic and personalized settings. Most approaches are built on U-Net~\cite{u-net} architectures. PBE~\cite{pbe} aligns visual and textual semantics for reference-guided completion, while AnyDoor~\cite{anydoor} enhances texture fidelity by injecting high-frequency features into U-Net, often at the expense of style consistency. Personalized methods~\cite{textual-inversion,e4t}, which learn instance-specific embeddings, improve identity preservation but require additional fine-tuning.

Recent virtual try-on frameworks increasingly adopt an inpainting-style formulation~\cite{dci-vton,cat-dm,ladi-vton,d4-vton}, treating garment transfer as conditional appearance completion under structural guidance. When extended to training-free virtual try-on, a central challenge is to enforce pose consistency without excessively restricting appearance synthesis. OmniVTON++ addresses this challenge by incorporating pose-aware guidance during diffusion sampling to regulate pose alignment while retaining appearance flexibility.

Beyond U-Net-based designs, DiT-based inpainting models have recently been actively explored~\cite{personalize,diptych}. Attention sharing strategies commonly adopted in U-Net-based methods are not inherently compatible with DiT architectures due to differences in attention formulation and positional encoding schemes. To support architectural generality across diffusion backbones, we resolve this incompatibility by enforcing distinguishable positional identities across sources, enabling reliable feature interaction across different model backbones.

\begin{figure*}
	\centering
	\includegraphics[width=\linewidth]{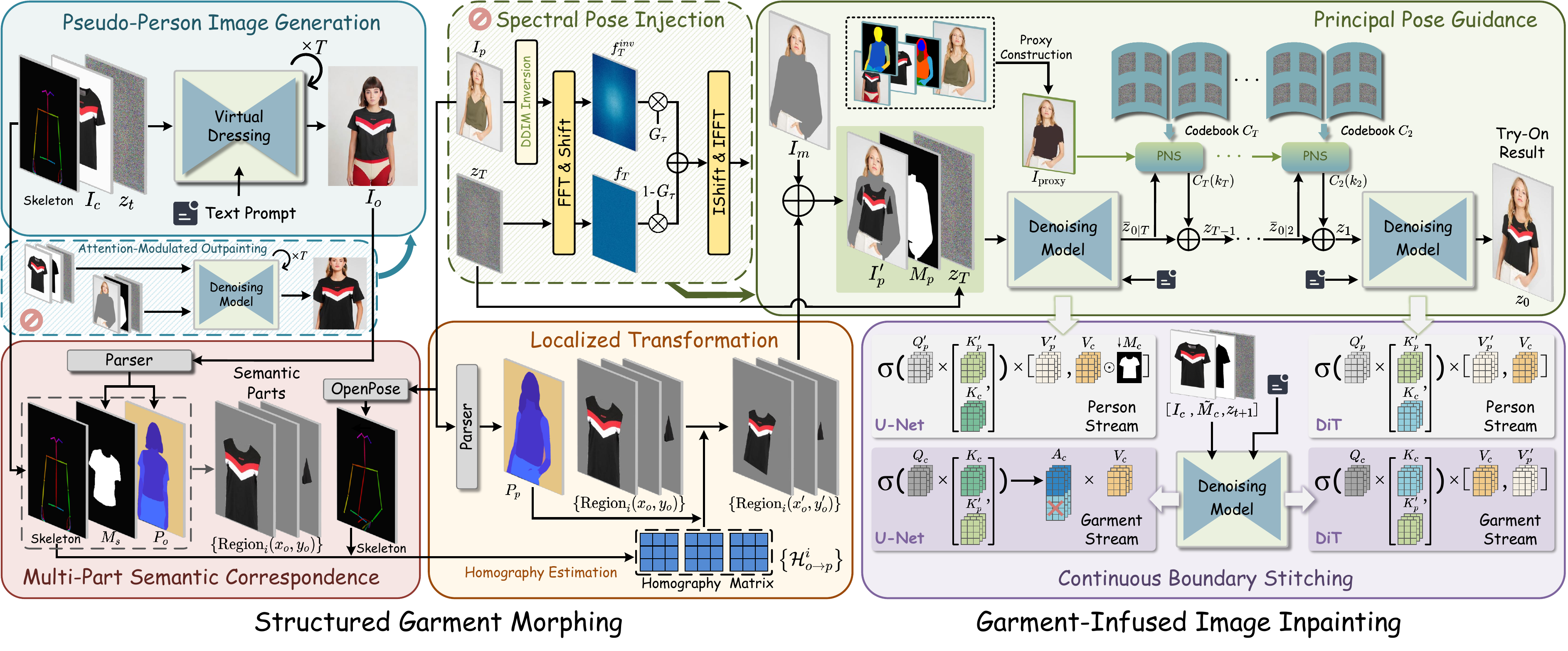}
	\captionof{figure}{Overview of OmniVTON++. OmniVTON++ performs Structured Garment Morphing (SGM) to obtain an adapted garment prior, which is incorporated into garment-infused image inpainting, where Principal Pose Guidance (PPG) enforces pose alignment and Continuous Boundary Stitching (CBS) improves boundary consistency across different model backbones. The crossed-circle marker (\(\oslash\)) indicates components derived from OmniVTON that are not included in the extended journal version.}
	\label{fig:pipeline}
\end{figure*}

\section{Approach}
\label{sec:method}
\subsection{Overview}
\textbf{OmniVTON}~\cite{omnivton} is a training-free virtual try-on framework that takes a garment image $I_c$ and a person image $I_p$ as input, where $I_c$ can be either a standalone flat-lay product or a garment worn by a person in arbitrary backgrounds. It addresses virtual try-on by explicitly establishing garment-to-body correspondence and directly computing geometry-aligned priors. The garment is first structured-morphed to match the body geometry of the person image, and the resulting body-aligned prior then guides garment-infused image inpainting together with pose-encoded noise. During inpainting, OmniVTON refines garment boundaries and completes the final appearance, producing coherent synthesis with consistent pose alignment. The framework demonstrates robustness to dataset shifts and garment type variations, and maintains effectiveness across diverse scenarios.

\noindent\textbf{OmniVTON++} extends this formulation by strengthening geometric initialization, enabling persistent pose control during sampling, and ensuring compatibility with diffusion transformer backbones. Within the Structured Garment Morphing stage, OmniVTON++ introduces a garment-centric virtual dressing step for in-shop garments to synthesize a target-garment-wearing person image, providing stable structural cues for correspondence estimation.

For pose control, OmniVTON++ introduces Principal Pose Guidance (PPG), which upgrades pose guidance from initialization-only conditioning to step-wise control during diffusion sampling. PPG constructs a pose-consistent proxy latent and aligns the principal components of each intermediate prediction during diffusion sampling with this proxy, steering the sampling process toward the intended human pose while allowing the remaining dimensions to retain flexibility for appearance conditioning.

For appearance refinement, OmniVTON++ maintains OmniVTON’s boundary-handling mechanism and adapts it for diffusion transformers via Positional Index Realignment, which eliminates positional ambiguity during attention computation without modifying model structure or retraining. As illustrated in Fig.~\ref{fig:pipeline}, these extensions integrate seamlessly into the OmniVTON workflow without altering its training-free formulation, expanding pose control capability and architectural compatibility.

\subsection{Structured Garment Morphing}
\label{sec:sgm}
We propose Structured Garment Morphing (SGM) to achieve appearance-consistent garment deformation under explicit garment-to-body correspondence. Unlike TPS-based methods~\cite{vitonhd} and flow-based methods~\cite{gp-vton}, which require retraining for different domains, SGM constrains garment morphing using skeletal information and parsing maps in a training-free manner. By jointly operating on a target-garment-wearing person image $I_o$ and the person image $I_p$, SGM derives a one-to-one garment-to-body assignment that maps each garment pixel to its anatomically corresponding region. 

While this mechanism is directly applicable in Non-Shop-to-X settings, where X denotes arbitrary person image domains, the Shop-to-X case lacks parseable human body structure because only an isolated garment image is available, causing keypoint detection and parsing to fail. To ensure universality, SGM synthesizes $I_o$ in this setting to provide stable structural cues for correspondence estimation.

\noindent\textbf{Pseudo-Person Image Generation.} 
We observed that text-driven image generation often fails to produce suitable pseudo-person images in this setting, due to limited control over body structure and sensitivity to prompt design. In the original OmniVTON, we therefore adopted an attention-modulated outpainting strategy that injects human semantics from $I_p$ into the pseudo-person synthesis process. While this design can yield visually plausible results, it remains influenced by the geometric properties of the flat-lay garment $I_c$. Such inputs exhibit uncontrolled variations in scale, perspective, and silhouette, which may be misinterpreted as cues about body shape or spatial arrangement. As a result, the synthesized pseudo-person images are not sufficiently reliable for part-based correspondence estimation and geometric mapping.

To support these downstream steps, we require a geometry-aware prior that provides a stable body layout and a reasonable worn form for the target garment. OmniVTON++ introduces this prior through a pretrained garment-centric virtual dressing module. By capturing garment–body geometry from real dressed-human images, its geometric estimation resolves the ambiguity inherent in flat-lay imagery and stabilizes the pseudo-person image generation process. Building on this capability, our pseudo-human synthesis takes $I_c$ as the primary condition and, together with keypoint constraints and textual prompts, generates an output with a plausible worn form of the target garment. In practice, we fix the pose to an arms-down standing configuration (A-pose) to avoid occlusion-induced ambiguities in part definition, from which the corresponding keypoints are extracted, and we employ a unified prompt template, such as ``white background, no occlusion,'' to ensure standardized outputs. The dressing module is used solely to obtain garment geometry and keypoints; therefore, any appearance bias in non-garment regions does not affect the geometric mapping stage of SGM.

\noindent\textbf{Multi-Part Semantic Correspondence.}
After obtaining a target-garment-wearing person image $I_o$, either directly available or synthesized when required, we establish part-level semantic correspondences between $I_o$ and $I_p$ using human skeletal cues. Keypoints are detected on both images with OpenPose~\cite{openpose} and grouped according to anatomical structure to define body parts.

Taking upper-body garments as a concrete example, we consider five parts: the torso, the left and right upper arms, and the left and right lower arms. Based on the detected keypoints, we construct bounding boxes $\{B_o^i\}_{i=1}^5$ around each part in $I_o$, with the corresponding boxes $\{B_p^i\}_{i=1}^5$ obtained in the same manner from $I_p$.

To derive pixel-level support within each part, we further introduce a human part segmentation map $P_o$, obtained from a human parsing model. Because such parsing does not distinguish garment pixels from surrounding non-garment content within a given part, each part mask $P_o^i$ is gated by the garment mask $M_o$ extracted from $I_o$, ensuring that only garment pixels are retained. The effective support of part $i$ is defined by the indicator function
\begin{equation}
\mathbb{I}_{\text{Region}_i}(x_o,y_o)=
\begin{cases}
1, & \text{if }(x_o,y_o)\in P_o^i \cap M_o \cap B_o^i,\\
0, & \text{otherwise.}
\end{cases}\label{eq:1}
\end{equation}
where $\mathbb{I}(\cdot)$ denotes the indicator function and $(x_o,y_o)$ represents pixel coordinates in $I_o$.

This construction yields a set of well-defined, garment-aware part supports in both $I_o$ and $I_p$, which serve as the basis for localized geometric mapping.

\noindent\textbf{Localized Transformations.} 
Given the established part supports, we estimate localized geometric mappings from $I_o$ to $I_p$. For each matched box pair $\{B^i_o,B^i_p\}$, a homography matrix $\mathcal{H}^i_{o\rightarrow p}\in\mathbb{R}^{3\times 3}$ is optimized directly using the Levenberg-Marquardt algorithm~\cite{levenberg}. 

These mappings are applied through piecewise perspective morphing to deform $I_o$ into the spatial configuration of $I_p$: 
\begin{equation}
\begin{bmatrix}
x_o'\\
y_o'\\
1
\end{bmatrix}
=\sum_{i} \mathbb{I}_{\text{Region}_i(x_o,y_o)}H_{o\rightarrow p}^i
\begin{bmatrix}
x_o\\
y_o\\1
\end{bmatrix},\label{eq:2}
\end{equation}
where $(x_o',y_o')$ denotes the resulting pixel coordinates. 

To regulate region composition while respecting occlusions induced by the person pose in $I_p$, we use the part segmentation map $P_p$ to retain, within each region, only pixels consistent with the corresponding part labels. This process yields a coarse morphed garment $I_w$, which serves as a prior for garment-infused image inpainting. Local discontinuities along part boundaries arising from this composition are handled by a boundary stitching mechanism described in Sec.~\ref{sec:cbs}.

\begin{figure*}[t]
	\centering
	\includegraphics[width=\linewidth]{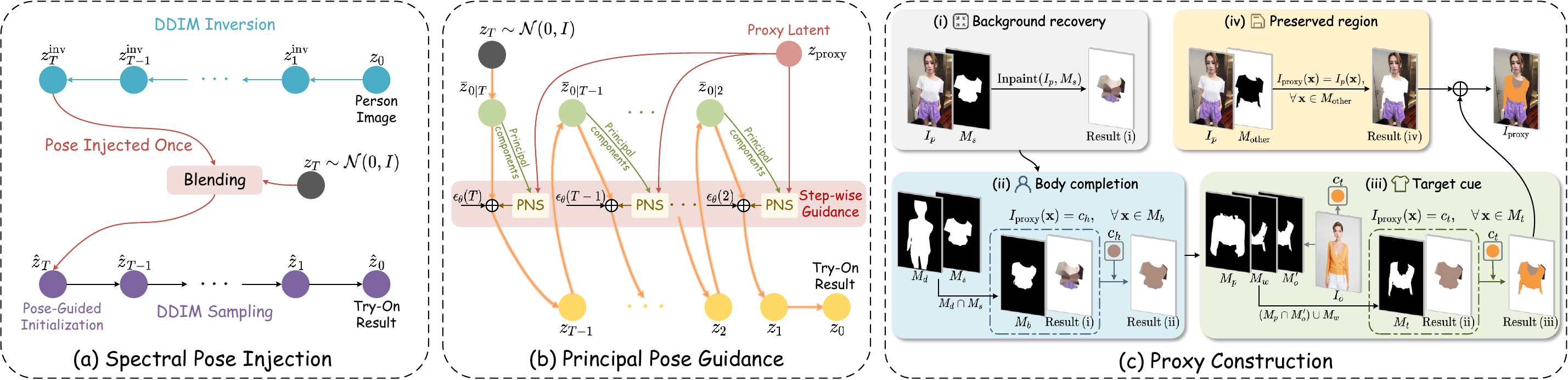}
	\captionof{figure}{Pose guidance mechanisms in OmniVTON and OmniVTON++. (a) Spectral Pose Injection (SPI) in OmniVTON incorporates human structure only at initialization by combining random noise with an inverted latent. (b) Principal Pose Guidance (PPG) in OmniVTON++ enforces pose-consistent structure throughout diffusion sampling via step-wise noise selection, guided by a proxy latent. (c) A proxy image is constructed via ordered region-wise composition to preserve pose while excluding garment appearance.}
	\label{fig:pca}
\end{figure*}

\subsection{Principal Pose Guidance}\label{sec:ppg}
Beyond garment deformation, virtual try-on also requires synthesizing plausible human structures within the inpainting region, where appearance is jointly determined by body pose and the morphed garment. Since this region typically over-approximates the human body, the synthesis must recover plausible body structure under substantial flexibility. Pose therefore provides a critical structural constraint.

Many existing approaches incorporate pose as an input conditioning signal. While effective, such designs rely on learned cross-modal interactions and task-specific training, limiting generalization across data and application settings as well as architectures. In response to this limitation, OmniVTON introduces Spectral Pose Injection (SPI) as an initialization-level mechanism for incorporating human structure by manipulating the initial noise prior to sampling (Fig.~\ref{fig:pca}(a)). Specifically, SPI combines a random noise $z_T$ with an inverted latent $z_T^{\mathrm{inv}}$ that preserves human structure, yielding a pose-guided initialization $\hat{z}_T$. 

OmniVTON++ extends this idea by introducing Principal Pose Guidance (PPG), which intervenes throughout the sampling process by constraining pose-consistent structure in intermediate predictions (Fig.~\ref{fig:pca}(b)). PPG is realized through a diffusion sampling framework with discrete noise selection, following the Discrete Diffusion Compression Model (DDCM) formulation~\cite{ddcm}. In this framework, diffusion sampling is guided by selecting noise based on the residual between a target latent $z_0$ and the intermediate prediction. The noise space at each timestep is discretized into timestep-specific codebooks $\{C_t\}$, and the latent state is updated as
\begin{equation}
z_{t-1} = \mu(z_t) + \sigma_t \epsilon_t, \qquad \epsilon_t \sim C_t ,
\end{equation}
under standard diffusion notation. At each timestep, the noise entry is selected by maximizing its alignment with the residual signal between the target latent and the intermediate prediction $\bar{z}_{0|t}$:
\begin{equation}
k_t =
\underset{k \in \{1,\ldots,K\}}{\arg\max}
\ \langle C_t(k),\, z_0 - \bar{z}_{0|t} \rangle ,
\end{equation}
where $K$ denotes the size of the codebook, and $\langle\cdot,\cdot\rangle$ denotes the inner product.

This step-wise noise selection provides a mechanism for progressively guiding the sampling process with respect to the human pose present in the person image. However, the residual-based selection in DDCM assumes access to the target latent $z_0$, which does not hold in virtual try-on, as $z_0$ corresponds to the synthesis objective itself. A straightforward alternative is to substitute the person-image latent for $z_0$, which enforces pose consistency but injects the original garment appearance into the guidance signal.

We therefore replace the unavailable latent $z_0$ with a proxy latent $z_{\text{proxy}}$, obtained by encoding a proxy image $I_{\text{proxy}}$. The proxy image preserves the human pose of $I_p$ while removing the original clothing appearance, and is constructed through an ordered composition process, in which each step overwrites previous values within its designated region (Fig.~\ref{fig:pca}(c)). 

\noindent\textbf{(i) Background recovery.}
We first recover background regions within the garment area of $I_p$ to remove garment appearance and restore background continuity. Pixels inside the garment mask are then completed using an image inpainting operator:
\begin{equation}
I_{\text{proxy}}(\mathbf{x}) = \operatorname{Inpaint}(I_p, M_s), \quad \forall\, \mathbf{x} \in M_s,\label{eq:5}
\end{equation}
where $M_s$ denotes the garment mask extracted from $I_p$.

\noindent\textbf{(ii) Body completion.}
Body completion is then performed to establish a coherent human-body support in areas previously covered by the original garment. We denote this region as
\begin{equation}
M_b = M_d \cap M_s ,\label{eq:6}
\end{equation}
where $M_d$ is the foreground human mask obtained from DensePose~\cite{densepose}. Pixels in $M_b$ are filled with a constant color:
\begin{equation}
I_{\text{proxy}}(\mathbf{x}) = c_h, \quad \forall\, \mathbf{x} \in M_b ,\label{eq:7}
\end{equation}
where $c_h$ represents the average skin color estimated from exposed body regions in $I_p$. This step establishes a clean body prior preserving pose structure.

\noindent\textbf{(iii) Target-garment cue.}
Target-garment regions are defined as
\begin{equation}
M_t = (M_p \cap M_o') \cup M_w ,\label{eq:8}
\end{equation}
where $M_p$ denotes the cloth-agnostic mask, $M_o'$ is the DensePose-projected target-garment mask warped onto the person image, and $M_w$ is a binary mask derived from the coarse morphed garment $I_w$. The union combines complementary properties for garment localization, where $M_o'$ provides pose-consistent placement but may be inaccurate along loose garment boundaries due to the body-surface parameterization of DensePose, while $M_w$ offers more stable shape coverage but may exhibit boundary discontinuities.

The target-garment region is then assigned a constant color,
\begin{equation}
I_{\text{proxy}}(\mathbf{x}) = c_t, \quad \forall\, \mathbf{x} \in M_t ,\label{eq:9}
\end{equation}
where $c_t$ is set to the average color of the target garment estimated from $I_o$, introducing a coarse target-garment-aware cue that conveys global statistics without texture detail.

\noindent\textbf{(iv) Preserved region.}
The preserved region is defined as
\begin{equation}
M_{\text{other}} = 1 - (M_s \cup M_b \cup M_t) .\label{eq:10}
\end{equation}
Within this region, original pixels from $I_p$ are retained:
\begin{equation}
I_{\text{proxy}}(\mathbf{x}) = I_p(\mathbf{x}), \quad \forall\, \mathbf{x} \in M_{\text{other}} .\label{eq:11}
\end{equation}

The resulting proxy image preserves pose information through its constant-valued composition. Using the proxy latent obtained from this image, PPG applies pose guidance to intermediate predictions during diffusion sampling. Empirically, we observe that the dominant components of the intermediate prediction $\bar{z}_{0|t}$ tends to capture pose-related structure, whereas the remaining components are primarily associated with fine-grained appearance variations. The pose guidance is therefore restricted to the principal structural components of $\bar{z}_{0|t}$, while leaving the remaining dimensions available for appearance modeling. This defines pose-guided noise selection (PNS) in PPG:
\begin{equation}
k_t =
\underset{k \in \{1,\ldots,K\}}{\arg\max}
\ \langle C_t(k),\, z_{\text{proxy}} - \bar{z}_{0|t}^{\text{PCA}} \rangle ,
\label{eq:12}
\end{equation}
where $\bar{z}_{0|t}^{\text{PCA}}$ denotes the principal components of $\bar{z}_{0|t}$, and $z_{\text{proxy}}$ is the proxy latent. The selected entry $C_t(k_t)$ is then incorporated into the diffusion update as
\begin{equation}
\begin{aligned}
z_{t-1}
&=
\sqrt{\alpha_{t-1}}\, \mu_\theta(z_t, t, c)
+
\sqrt{1 - \alpha_{t-1} - \sigma_t^2}\, \epsilon_\theta(z_t, t, c) \\
&\quad
+
\sigma_t\, C_t(k_t),
\end{aligned}\label{eq:13}
\end{equation}
where $\epsilon_\theta(\cdot)$ denotes the noise prediction network, $c = [I_p'; M_p; c_{\text{txt}}]$ represents the conditioning inputs, $I_p'$ is the garment-infused person image, and $c_{\text{txt}}$ is the text prompt describing the target garment. By injecting pose-guided noise, the sampling process is constrained to maintain human body structure, while garment appearance remains governed by other conditioning signals.

\subsection{Continuous Boundary Stitching}       
\label{sec:cbs}
SGM enables explicit, training-free garment deformation via correspondence-based region composition; however, assembling the coarse garment prior $I_w$ in this manner can introduce boundary discontinuities at region interfaces. During diffusion-based appearance completion, these discontinuities may be absorbed as garment structures, yielding spurious seams or pattern misalignment in the try-on results. 

To mitigate this issue, we propose Continuous Boundary Stitching (CBS), a training-free boundary refinement mechanism based on a dual-stream design. CBS treats the original try-on generation pathway as the person stream, and introduces an additional garment stream to provide complementary garment-side information. Both streams are initialized from the same diffusion noise and maintain separate latent representations throughout sampling. The person stream takes the garment-infused person image $I_p'$ and the cloth-agnostic mask $M_p$, whereas the garment stream incorporates the garment image $I_c$ together with the inverted cloth mask $\tilde{M}_c$. Accordingly, CBS modifies the self-attention computation to enable bidirectional cross-stream feature interaction during sampling.

We first describe the person stream, where attention jointly attends to person-stream and garment-stream features. Let $Q_p'$, $K_p'$, $V_p'$ denote the query, key, and value features of the person stream, and $Q_c$, $K_c$, $V_c$ denote the corresponding features of the garment stream. The person-stream feature $f_p'$ is computed as
\begin{equation}
f_p' =
\operatorname{Softmax}\!\left(
\frac{Q_p' \cdot [K_p' \parallel K_c]^\top}{\sqrt{d}}
\right)
\big[ V_p' \parallel (V_c \cdot D(M_c)) \big],\label{eq:14}
\end{equation}
where $d$ denotes the feature dimension, and $D(\cdot)$ downsamples the cloth mask $M_c$ to match the spatial resolution of $V_c$. This formulation allows person queries to leverage spatially continuous garment features to compensate for discontinuities introduced by region-wise morphing, while the masked garment values restrict cross-stream contributions to cloth regions and prevent background features from interfering with boundary reasoning.

In the garment stream, attention is guided by person-stream structure while preserving garment-side feature continuity. The garment-stream attention map $A_c$ is computed as
\begin{equation}
A_c =
\operatorname{Softmax}\!\left(
\frac{Q_c \cdot [K_c \parallel K_p']^\top}{\sqrt{d}}
\right).
\label{eqn:cstream}
\end{equation}
Since attention weights are jointly normalized over both garment and person key sets, the garment-stream attention distribution can adapt to the spatial layout implied by $I_p'$. To prevent texture interference from discontinuous person features, we exclude $V_p'$ from the output computation and aggregate only garment values:
\begin{equation}
f_c = A_c[:, 1\!:\!n] \cdot V_c,\label{eq:16}
\end{equation}
where $A_c \in \mathbb{R}^{n \times 2n}$, $n$ denotes the sequence length of the flattened spatial features, and $A_c[:, 1\!:\!n]$ corresponds to the submatrix associated with the $K_c$ block after joint normalization. In this way, the garment stream results in internally continuous, layout-guided garment features that provide boundary cues to the person stream over successive denoising steps.

\subsection{Positional Index Realignment}
\label{sec:dit}
\begin{figure}
	\centering
	\includegraphics[width=\linewidth]{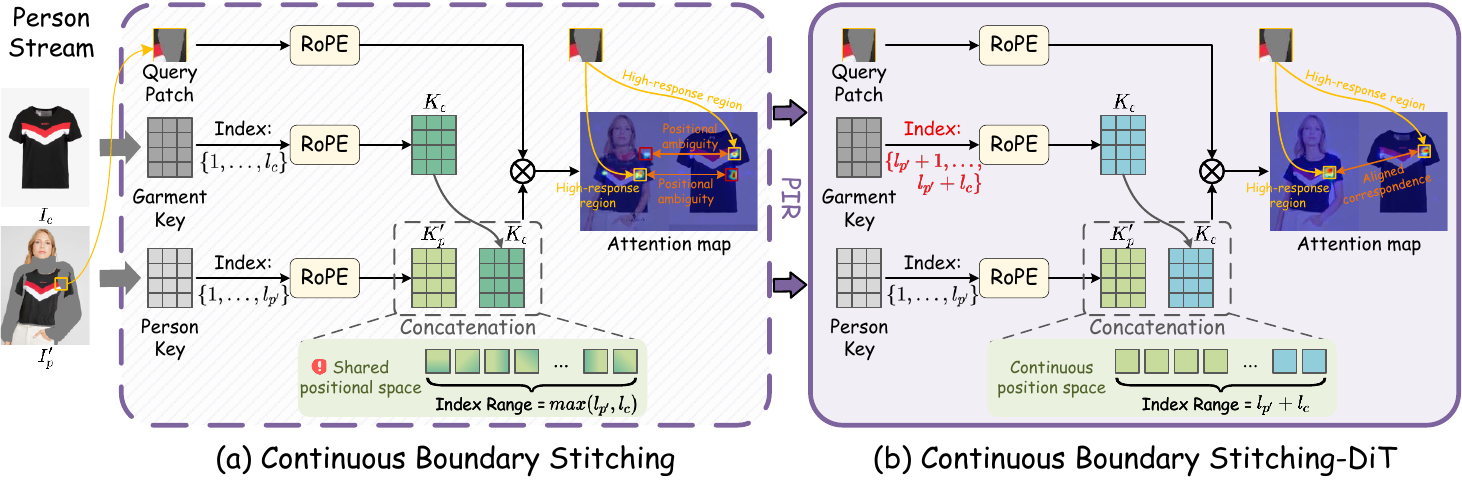}
	\captionof{figure}{Cross-stream correspondence in CBS and CBS-DiT on DiT backbones. Person-stream attention is shown as an illustrative example. (a) CBS exhibits positional ambiguity under a shared positional space. (b) CBS-DiT with Positional Index Realignment (PIR) establishes a continuous positional space, yielding aligned correspondence.}
	\label{fig:attn_vis}
\end{figure} 

CBS remains robust under U-Net-based diffusion backbones, yet its direct application to Diffusion Transformers (DiTs) such as FLUX~\cite{flux} leads to marked degradation. This issue arises in DiT backbones when tokens from different image inputs are processed within a shared attention sequence under a common rotary positional encoding (RoPE)~\cite{roformer} index domain. In this setting, tokens originating from different image inputs may receive identical positional encodings despite representing distinct semantic regions. With attention-level interaction across streams in CBS, such positional ambiguity can lead to misaligned correspondences, as illustrated in Fig.~\ref{fig:attn_vis}(a), and may further result in visible artifacts in the generated outputs.

To resolve this issue, we propose Positional Index Realignment (PIR), a minimal and training-free adaptation for DiT backbones. PIR operates at the level of positional index assignment, mapping tokens from different inputs to non-overlapping index ranges. Let the token lengths of the text, person-image, and garment-image inputs be denoted as $l_{\text{txt}}, l_{p'}, l_c$, respectively, and define the total sequence length as
\begin{equation}
l = l_{\text{txt}} + l_{p'} + l_c .
\end{equation}
We then precompute the rotary bases up to length $l$, with
\begin{equation}
\cos, \sin \in \mathbb{R}^{l \times d},
\end{equation}
where $d$ denotes the per-head feature dimension used by RoPE. During encoding, text tokens occupy the initial index segment, while person-image and garment-image tokens are assigned to subsequent disjoint segments following the text portion. This realignment preserves the internal geometric structure of each image input while preventing positional collisions across inputs (Fig.~\ref{fig:attn_vis}(b)).

With positional indices realigned, CBS can be instantiated on DiT backbones. We refer to this setting as CBS-DiT, with feature aggregation defined as:
\begin{equation}\small
\begin{alignedat}{2}
f_p^\prime
&= \operatorname{Softmax}\!\left(
   \frac{
     Q_{p}^\prime \cdot
     [K_{p}^\prime \parallel K_{c}[:,\, l_{\text{txt}}+1:l_{\text{txt}}+l_c]]^\top
   }{\sqrt{d}}
 \right) \\
&\quad \times
 [V_{p}^\prime \parallel V_{c}[:,\, l_{\text{txt}}+1:l_{\text{txt}}+l_c]], \\[3pt]
f_c
&= \operatorname{Softmax}\!\left(
   \frac{
     Q_{c} \cdot
     [K_{c} \parallel K_{p}^\prime[:,\, l_{\text{txt}}+1:l_{\text{txt}}+l_{p'}]]^\top
   }{\sqrt{d}}
 \right) \\
&\quad \times
 [V_{c} \parallel V_{p}^\prime[:,\, l_{\text{txt}}+1:l_{\text{txt}}+l_{p'}]] .
\end{alignedat}
\label{eq:19}
\end{equation}

Different from the U-Net-based implementation, we avoid mask-based modulation in the person stream and do not suppress person-stream values in the garment stream. In DiT backbones, attention is normalized jointly across text and image tokens; under this normalization, such interventions disrupt token alignment and degrade boundary coherence. We therefore retain full attention aggregation in CBS-DiT.

\section{Experiments}
\label{sec:exp}
\subsection{Experimental Setup}
\noindent\textbf{Implementation Details.}
All experiments are implemented in PyTorch 2.5.1 and conducted on a single NVIDIA GeForce RTX A6000 GPU, with OmniVTON++ instantiated on top of both Stable Diffusion v2.0 (SD-2.0)~\cite{sd} and FLUX.1 Fill~\cite{flux}.

For SGM, we use IMAGDressing-v1~\cite{imagdressing} with default settings to synthesize the pseudo-person image required for geometry alignment. Region-level processing is applied according to garment categories: 1) Upper-body garments are decomposed into five regions, including the left and right upper arms, left and right lower arms, and the torso;
2) Lower-body garments are partitioned into five regions corresponding to the left and right upper legs, left and right lower legs, and the hip-above region; 3) Dresses are separated into upper- and lower-body sections. We obtain human parsing maps $P_p$ and $P_o$ using TAPPS~\cite{TAPPS}, which provides fine-grained human part segmentation but does not distinguish garment regions; instance-level parser PGN~\cite{pgn} is therefore employed to extract garment masks $M_o$ and $M_s$. For benchmark datasets, agnostic masks and clothing masks are directly provided. In scenarios without available annotations, general-purpose segmentation tools such as SAM~\cite{sam} can be used to obtain the corresponding masks from user images. In the background recovery step of proxy construction, we use a Navier-Stokes-based image inpainting approach~\cite{inpaint_ns}.

For the garment-infused image inpainting stage, we use the DDIM sampler with 50 steps for SD-2.0, and the SDE variant of DPM-Solver++~\cite{dpm-solver++} with 30 steps for FLUX. PPG is implemented by replacing the stochastic noise term via PNS while keeping the sampling update unchanged, where 3 principal components are used at each timestep. The codebook size $K$ is fixed to 64. Text prompts are automatically generated using CLIP Interrogator~\cite{clipin}. The overall workflow of OmniVTON++ is summarized in Algorithm~\ref{alg:omnivtonpp}.

\begin{algorithm}[t]
\caption{OmniVTON++}
\label{alg:omnivtonpp}
\KwIn{\begin{tabular}[t]{@{}l@{}}
person image $I_p$; garment image $I_c$;\\
text prompt $c_{txt}$; person mask $M_p$, \\
garment masks $M_o$, $M_s$, body mask $M_d$;\\
diffusion steps $T$; backbone type (U-Net or DiT).
\end{tabular}}
\KwOut{try-on result.}
\tcp{SGM}
synthesize a target-garment-wearing image $I_o$ for the Shop-to-X setting; otherwise set $I_o \leftarrow I_c$\;
obtain parsing maps $P_p$ and $P_o$\;
estimate part-level correspondences (Eq.~(\ref{eq:1}))\;
perform localized morphing (Eq.~(\ref{eq:2}))\;
compose coarse morphed garment $I_w$ respecting $P_p$\;
\tcp{Proxy construction for PPG}
\Indp
(i) background recovery (Eq.~(\ref{eq:5}))\;
(ii) body completion (Eqs.~(\ref{eq:6})--(\ref{eq:7}))\;
(iii) inject target-garment cue (Eqs.~(\ref{eq:8})--(\ref{eq:9}))\;
(iv) preserve remaining regions (Eqs.~(\ref{eq:10})--(\ref{eq:11}))\;
\Indm
\If{DiT backbone}{apply PIR and set CBS-DiT mode (Eq.~(\ref{eq:19}))\;}
\If{U-Net backbone}{set CBS mode (Eqs.~(\ref{eq:14})--(\ref{eq:16}))\;}
\tcp{Sampling with PPG and CBS}
encode $I_{\text{proxy}}$ to obtain $z_{\text{proxy}}$\;
set conditioning $c=[I_p';\,M_p;\,c_{txt}]$ and initialize $z_T$\;
\For{$t=T$ \KwTo $1$}{
    predict $\bar{z}_{0|t}$ from $z_t$\;
    perform pose-guided noise selection (Eq.~(\ref{eq:12}))\;
    update $z_t \rightarrow z_{t-1}$ (Eq.~(\ref{eq:13}))\;
}
decode $z_0$ to obtain the try-on result\;
\end{algorithm}

\begin{table}[!h]\centering
\def\arraystretch{1.2}
\small
\scriptsize
\tabcolsep 1.5pt
\captionof{table}{Quantitative comparisons on VITON-HD under the Shop-to-Model setting. Subscripts $u$ and $p$ denote the unpaired and paired settings, respectively. The \colorbox[RGB]{255,179,179}{best}, \colorbox[RGB]{255,217,179}{second-best}, and \colorbox[RGB]{255,255,179}{third-best} results are highlighted in each cell.}
\resizebox{\linewidth}{!}{
    \begin{tabular}{l cc ccccc}
        \toprule
        Method && Venue \& Year && FID$_u$$\downarrow$ & FID$_p$$\downarrow$ & SSIM$_p$$\uparrow$ & LPIPS$_p$$\downarrow$ \\
        \cmidrule{1-1} \cmidrule{3-3} \cmidrule{5-8}
        \rowcolor[HTML]{F2F2F2}
        \multicolumn{8}{l}{\textit{Virtual Try-on Method}} \\
        GP-VTON~\cite{gp-vton} && CVPR 2023 && 51.566 & 49.196 & 0.810 & 0.249 \\
        CAT-DM~\cite{cat-dm} && CVPR 2024 && 28.869 & 26.339 & 0.775 & 0.229 \\
        D$^4$-VTON~\cite{d4-vton} && ECCV 2024 && 25.299 & 23.914 & 0.790 & 0.250 \\
        IDM-VTON~\cite{idm-vton} && ECCV 2024 && 23.035 & 20.460 & 0.812 & 0.147 \\
        OOTDiffusion~\cite{ootd} && AAAI 2025 && 26.312 & 21.560 & 0.757 & 0.208 \\       
        Any2AnyTryon~\cite{any2anytryon} && ICCV 2025 && 18.225 & 16.742 & 0.645 & 0.365 \\
        \cmidrule{1-1} \cmidrule{3-3} \cmidrule{5-8}
        \rowcolor[HTML]{F2F2F2}
        \multicolumn{8}{l}{\textit{Exemplar-based Image Editing Method}} \\
        PBE~\cite{pbe} && CVPR 2023 && 19.230 & 17.649 & 0.784 & 0.227 \\
        AnyDoor~\cite{anydoor} && CVPR 2024 && 14.830 & 9.922 & 0.796 & 0.164 \\
        TIGIC~\cite{tigic} && ECCV 2024 && 90.338 & 88.900 & 0.613 & 0.422 \\
        Cross-Image~\cite{cross-image} && SIGGRAPH 2024 && 62.614 & 57.286 & 0.760 & 0.256 \\
        \cmidrule{1-1} \cmidrule{3-3} \cmidrule{5-8}
        \rowcolor[HTML]{F2F2F2}
        \multicolumn{8}{l}{\textbf{\textit{Our Method}}} \\
        OmniVTON~\cite{omnivton} && ICCV 2025 && \third{9.621} & \third{7.758} & \third{0.832} & \third{0.145} \\        
        OmniVTON++ (SD-2.0) && - && \best{9.189} & \second{6.990} & \second{0.843} & \second{0.130} \\
        OmniVTON++ (FLUX) && - && \second{9.286} & \best{6.618} & \best{0.849} & \best{0.121} \\
        \bottomrule
    \end{tabular}
}
\label{tab:vitonhd}
\end{table}

\noindent\textbf{Datasets.}
We evaluate OmniVTON++ on two in-shop datasets, VITON-HD~\cite{vitonhd} and DressCode~\cite{dresscode}, as well as one in-the-wild dataset, DeepFashion2~\cite{deepfashion2}. VITON-HD provides 2,032 test pairs of upper garments and person images, while DressCode covers three garment categories (upper, lower, and dresses) with a total of 5,400 test samples. For DeepFashion2, following the StreetTryOn benchmark~\cite{street-tryon}, we construct a test set of 2,089 images spanning four try-on scenarios: Shop-to-Street, Model-to-Model, Model-to-Street, and Street-to-Street. The input resolution is set according to the person image, using $512\times384$ for VITON-HD and DressCode, and $512\times320$ for DeepFashion2.

\noindent\textbf{Baselines and Metrics.}
We compare OmniVTON++ against two categories of baselines. The first category consists of state-of-the-art virtual try-on methods, including GP-VTON~\cite{gp-vton}, CAT-DM~\cite{cat-dm}, D$^4$-VTON~\cite{d4-vton}, IDM-VTON~\cite{idm-vton}, OOTDiffusion~\cite{ootd}, PWS~\cite{pws}, PastaGAN++~\cite{pastagan++}, StreetTryOn~\cite{street-tryon}, and Any2AnyTryOn~\cite{any2anytryon}. Most of these baselines focus on the Shop-to-Model scenario. PWS and PastaGAN++ specifically support Model-to-X settings. StreetTryOn and Any2AnyTryOn are applicable to both in-shop and in-the-wild scenarios. The second category includes exemplar-based image editing methods: PBE~\cite{pbe}, AnyDoor~\cite{anydoor}, TIGIC~\cite{tigic}, and Cross-Image~\cite{cross-image}. These methods are not originally designed for virtual try-on but are included as complementary baselines, as they support reference-based image editing without retraining. All exemplar-based editing methods are evaluated using their official implementations with default settings.

Evaluation follows standard protocols. We use Fr\'echet Inception Distance (FID)~\cite{fid} to measure the similarity between generated try-on results and real image distributions. For VITON-HD and DressCode, which provide paired ground-truth images, we also report Structural Similarity (SSIM)~\cite{ssim} and Learned Perceptual Image Patch Similarity (LPIPS)~\cite{lpips} to assess structural consistency and perceptual quality, respectively.

\begin{table*}
  \centering
  \def\arraystretch{1.2}
  \scriptsize
  \tabcolsep 1.5pt
  \captionof{table}{Quantitative comparisons on DressCode under the Shop-to-Model setting. Subscripts $u$ and $p$ denote the unpaired and paired settings, respectively. The \colorbox[RGB]{255,179,179}{best}, \colorbox[RGB]{255,217,179}{second-best}, and \colorbox[RGB]{255,255,179}{third-best} results are highlighted in each cell.}
  \resizebox{\textwidth}{!}{
    \begin{tabular}{l cc ccccccccccccccc}
      \toprule
      \multirow{2}{*}{Method} && \multirow{2}{*}{Venue \& Year}
      && \multicolumn{4}{c}{DressCode-Upper}
      && \multicolumn{4}{c}{DressCode-Lower}
      && \multicolumn{4}{c}{DressCode-Dresses} \\
      \cmidrule{5-8} \cmidrule{10-13} \cmidrule{15-18}
      &&
      && FID$_u$$\downarrow$ & FID$_p$$\downarrow$ & SSIM$_p$$\uparrow$ & LPIPS$_p$$\downarrow$
      && FID$_u$$\downarrow$ & FID$_p$$\downarrow$ & SSIM$_p$$\uparrow$ & LPIPS$_p$$\downarrow$
      && FID$_u$$\downarrow$ & FID$_p$$\downarrow$ & SSIM$_p$$\uparrow$ & LPIPS$_p$$\downarrow$ \\
      \cmidrule{1-1} \cmidrule{3-4} \cmidrule{5-8} \cmidrule{10-13} \cmidrule{15-18}
      \rowcolor[HTML]{F2F2F2}
      \multicolumn{18}{l}{\textit{Virtual Try-On Method}} \\
      GP-VTON~\cite{gp-vton} && CVPR 2023
      && 50.628 & 47.234 & 0.879 & 0.183
      && 102.018 & 99.136 & 0.886 & 0.195
      && 65.028 & 63.381 & 0.770 & 0.274 \\
      CAT-DM~\cite{cat-dm} && CVPR 2024
      && 14.772 & 12.362 & 0.904 & 0.077
      && 21.990 & 18.366 & 0.870 & 0.116
      && 34.610 & 32.558 & 0.800 & 0.181 \\
      D$^4$-VTON~\cite{d4-vton} && ECCV 2024
      && 20.726 & 18.243 & 0.765 & 0.276
      && 34.088 & 31.067 & 0.746 & 0.284
      && 42.230 & 41.418 & 0.764 & 0.262 \\
      IDM-VTON~\cite{idm-vton} && ECCV 2024
      && 14.385 & 12.164 & 0.895 & 0.083
      && 21.554 & 17.777 & 0.855 & 0.124
      && 19.745 & 17.490 & 0.775 & 0.207 \\
      OOTDiffusion~\cite{ootd} && AAAI 2025
      && 25.840 & 13.424 & 0.879 & 0.089
      && 43.288 & 21.251 & 0.849 & 0.120
      && 52.229 & 24.609 & 0.787 & 0.175 \\
      Any2AnyTryon~\cite{any2anytryon} && ICCV 2025
      && 50.142 & 50.655 & 0.655 & 0.451
      && 25.580 & 23.270 & 0.691 & 0.361 
      && 32.762 & 26.601 & 0.705 & 0.335 \\
      \cmidrule{1-1} \cmidrule{3-4} \cmidrule{5-8} \cmidrule{10-13} \cmidrule{15-18}
      \rowcolor[HTML]{F2F2F2}
      \multicolumn{18}{l}{\textit{Exemplar-based Image Editing Method}} \\
      PBE~\cite{pbe} && CVPR 2023
      && 18.010 & 17.405 & 0.885 & 0.110
      && 20.394 & 17.996 & 0.866 & 0.129
      && 35.857 & 33.599 & 0.785 & 0.219 \\
      AnyDoor~\cite{anydoor} && CVPR 2024
      && 25.432 & 23.941 & 0.856 & 0.140
      && 23.338 & 21.670 & 0.811 & 0.180
      && 33.440 & 33.497 & 0.727 & 0.286 \\
      TIGIC~\cite{tigic} && ECCV 2024
      && 65.407 & 62.589 & 0.844 & 0.214
      && 78.359 & 79.137 & 0.779 & 0.284
      && 113.062 & 114.963 & 0.625 & 0.459 \\
      Cross-Image~\cite{cross-image} && SIGGRAPH 2024
      && 46.309 & 44.111 & 0.865 & 0.140
      && 42.674 & 37.486 & 0.847 & 0.141
      && 76.940 & 68.095 & 0.811 & 0.201 \\
      \cmidrule{1-1} \cmidrule{3-4} \cmidrule{5-8} \cmidrule{10-13} \cmidrule{15-18}
      \rowcolor[HTML]{F2F2F2}
      \multicolumn{18}{l}{\textbf{\textit{Our Method}}} \\
      OmniVTON~\cite{omnivton} && ICCV 2025
      && \third{12.917} & \third{11.520} & \third{0.899} & \third{0.087}
      && \third{14.992} & \third{12.678} & \third{0.873} & \third{0.110}
      && \second{15.520} & \second{13.938} & \third{0.798} & \third{0.174} \\  
      OmniVTON++ (SD-2.0) && -
      && \second{12.425} & \second{9.766} & \second{0.911} & \second{0.075}
      && \best{14.290} & \second{11.707} & \second{0.890} & \second{0.100}
      && \best{15.159} & \best{13.474} & \second{0.819} & \second{0.163} \\ 
      OmniVTON++ (FLUX) && -
      && \best{11.478} & \best{8.556} & \best{0.918} & \best{0.065}
      && \second{14.281} & \best{10.935} & \best{0.893} & \best{0.098}
      && \third{17.632} & \third{14.615} & \best{0.821} & \best{0.158} \\
      \bottomrule
    \end{tabular}
  }
  \label{tab:dresscode}
\end{table*}

\begin{table}[t]
\centering
\def\arraystretch{1.2}
\normalsize
\tabcolsep 1.5pt
\captionof{table}{Quantitative comparisons on the StreetTryOn benchmark. The \colorbox[RGB]{255,179,179}{best}, \colorbox[RGB]{255,217,179}{second-best}, and \colorbox[RGB]{255,255,179}{third-best} results are highlighted in each cell.
}
\resizebox{\linewidth}{!}{
\begin{tabular}{lc cc cc cc c}
\toprule
&& Shop-to-Street && Model-to-Model && Model-to-Street && Street-to-Street \\
\cmidrule{3-3} \cmidrule{5-5} \cmidrule{7-7} \cmidrule{9-9}
&& FID$\downarrow$ && FID$\downarrow$ && FID$\downarrow$ && FID$\downarrow$ \\
\cmidrule{1-1} \cmidrule{3-3} \cmidrule{5-5} \cmidrule{7-7} \cmidrule{9-9}
\rowcolor[HTML]{F2F2F2}
\multicolumn{9}{l}{\textit{Virtual Try-On Method}} \\
CAT-DM~~\cite{cat-dm} && 37.484 && - && - && - \\ 
D$^4$-VTON~~\cite{d4-vton} && 35.003 && - && - && - \\ 
IDM-VTON~~\cite{idm-vton} && 42.282 && - && - && - \\ 
OOTDiffusion~\cite{ootd} && 42.318 && - && - && - \\
PWS~~\cite{pws} && - && 34.858 && 77.274 && 84.990 \\
PastaGAN++~~\cite{pastagan++} && - && 13.848 && 71.090 && 67.016 \\
StreetTryOn~~\cite{street-tryon} && 34.054 && 12.185 && 34.191 && 33.039 \\
Any2AnyTryon~~\cite{any2anytryon} && 98.117 && 22.533 && 96.750 && 64.833 \\
\cmidrule{1-1} \cmidrule{3-3} \cmidrule{5-5} \cmidrule{7-7} \cmidrule{9-9}
\rowcolor[HTML]{F2F2F2}
\multicolumn{9}{l}{\textit{Exemplar-based Image Editing Method}} \\
PBE~~\cite{pbe} && 81.538 && 20.181 && 62.664 && 36.556 \\ 
AnyDoor~~\cite{anydoor} && 50.893 && 24.235 && 51.861 && 35.139 \\ 
TIGIC~~\cite{tigic} && 100.177&& 114.151 && 130.836 && 121.520 \\ 
Cross-Image~~\cite{cross-image} && 69.444 && 52.310 && 66.755 && 57.753 \\
\cmidrule{1-1} \cmidrule{3-3} \cmidrule{5-5} \cmidrule{7-7} \cmidrule{9-9}
\rowcolor[HTML]{F2F2F2}
\multicolumn{9}{l}{\textbf{\textit{Our Method}}} \\
OmniVTON~\cite{omnivton} && \third{33.919} && \second{8.983} && \third{33.450} && \third{23.470} \\ 
OmniVTON++ (SD-2.0) && \second{32.357} && \best{8.772} && \second{32.894} && \second{21.960} \\
OmniVTON++ (FLUX) && \best{32.150} && \third{9.096} && \best{32.479} && \best{21.272} \\
\bottomrule
\end{tabular}}
\label{tab:street_tryon}
\end{table}

\subsection{Comparison with State-of-the-Art Methods}
\noindent\textbf{Quantitative Evaluation.}
We first evaluate the \textit{\textbf{cross-dataset}} generalization of OmniVTON++, where all VTON competitors are tested on VITON-HD using their official checkpoints pre-trained on DressCode, with Any2AnyTryOn evaluated using its general checkpoint. Quantitative results are reported in Tab.~\ref{tab:vitonhd}. In particular, we report OmniVTON++ instantiated with two diffusion models, SD-2.0 and FLUX, enabling comparison \textit{\textbf{across backbone architectures}}.

Both variants rank first or second on all reported metrics. Specifically, OmniVTON++ built on SD-2.0 achieves the lowest FID under the unpaired setting, while the FLUX-based variant attains the best results on the remaining metrics. This complementary performance may be attributed to differences in architectural design and pre-training data priors. Among the baselines, AnyDoor yields the best FID, likely benefiting from training data that include VITON-HD samples. However, its structural accuracy and perceptual quality remain inferior to those of OmniVTON++, as it is designed for general image editing without mechanisms tailored for garment conditioning. In contrast, the VTON method IDM-VTON better preserves garment structure yet shows weaker generalization compared to OmniVTON++. Note that OmniVTON++ also improves over OmniVTON in this evaluation.

We further evaluate \textit{\textbf{cross-garment-type}} generalization by testing VTON methods pre-trained on VITON-HD (which contains only upper garments) on the DressCode dataset, which covers a wider range of clothing types including Upper, Lower, and Dresses. As reported in Tab.~\ref{tab:dresscode}, the two OmniVTON++ variants occupy the top two positions on most reported metrics under this setting. The FLUX variant ranks first on all metrics for the Upper category and maintains top-tier performance for Lower and Dresses, while the SD-2.0 variant ranks within the top two across all metrics for all categories. Compared to OmniVTON, OmniVTON++ benefits from a refined garment prior and improved pose alignment, which contribute to the observed performance gains.

To evaluate generalization \textit{\textbf{across scenarios}}, we consider four additional settings beyond Shop-to-Model and summarize the results on the StreetTryOn benchmark in Tab.~\ref{tab:street_tryon}. Missing entries (“--”) indicate scenario-specific limitations of individual methods. No additional training is performed on the StreetTryOn dataset, as the goal is to assess generalization rather than in-domain performance. Specifically, most virtual try-on methods are evaluated using checkpoints trained on VITON-HD, which matches the upper-garment-only Shop-to-Street setting in StreetTryOn. PWS and PastaGAN++ are evaluated using official checkpoints trained on DeepFashion~\cite{deepfashion} and UPT~\cite{pastagan}, respectively. Results for StreetTryOn are taken from the original paper due to the absence of a public implementation, while GP-VTON is excluded from this comparison, as it relies on a task-specific parsing representation whose generating model is not released.

Under the Shop-to-Street setting, D$^4$-VTON and StreetTryOn incorporate explicit warping priors and outperform prior-free approaches. OmniVTON++ further enhances pose fidelity through step-wise control, leading to improved reconstruction quality. Across the remaining settings, our training-free framework delivers the best performance among the few applicable competitors, notably exceeding StreetTryOn despite the latter being trained on in-domain data. Moreover, as StreetTryOn relies on garment DensePose~\cite{garment-densepose}, which does not support lower-body garments or dresses, it cannot be applied to Shop-to-X scenarios involving these clothing categories. By contrast, SGM supports deformation of arbitrary garment types, enabling consistent performance across varied scenarios. Both variants of OmniVTON++ improve over OmniVTON in nearly all scenario transitions, with the Model-to-Model case under the FLUX instantiation being the only exception.

\begin{figure*}[t]
	\centering
	\includegraphics[width=\linewidth]{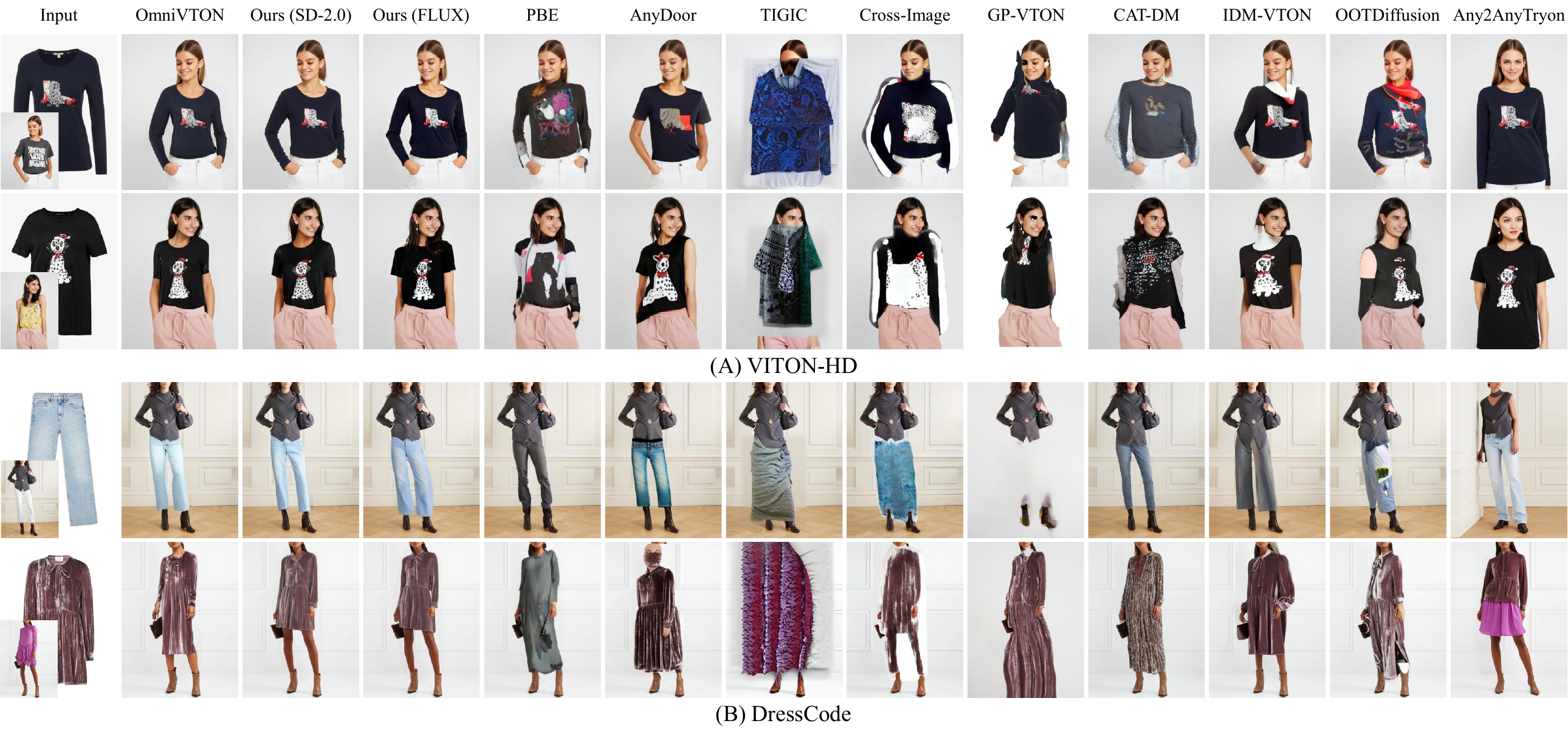}
	\captionof{figure}{Qualitative results under cross-dataset and cross-garment-type evaluation. Upper-garment try-on results on VITON-HD are shown on the top, while lower-garment and dress results on DressCode are shown on the bottom.}
	\label{fig:vitonhd_dresscode}
\end{figure*}

\noindent\textbf{Qualitative Evaluation.}  
We present qualitative results on the VITON-HD and DressCode datasets in Fig.~\ref{fig:vitonhd_dresscode}. Most traditional VTON methods, together with exemplar-based editing approaches such as PBE and AnyDoor, produce visually reasonable human-body structures, but often fail to transfer garment textures accurately and introduce noticeable artifacts in cross-dataset and cross-garment-type settings. As a GAN-based method, GP-VTON suffers from incomplete human-body synthesis, reflecting its limited generalization capability outside the training distribution. TIGIC and Cross-Image are unable to generate realistic human images due to the lack of coherent garment–body composition. In contrast, OmniVTON++ produces the most visually convincing try-on results, yielding more plausible garment geometry and more faithful human pose preservation than OmniVTON.

\begin{figure}
	\centering
	\includegraphics[width=0.95\linewidth]{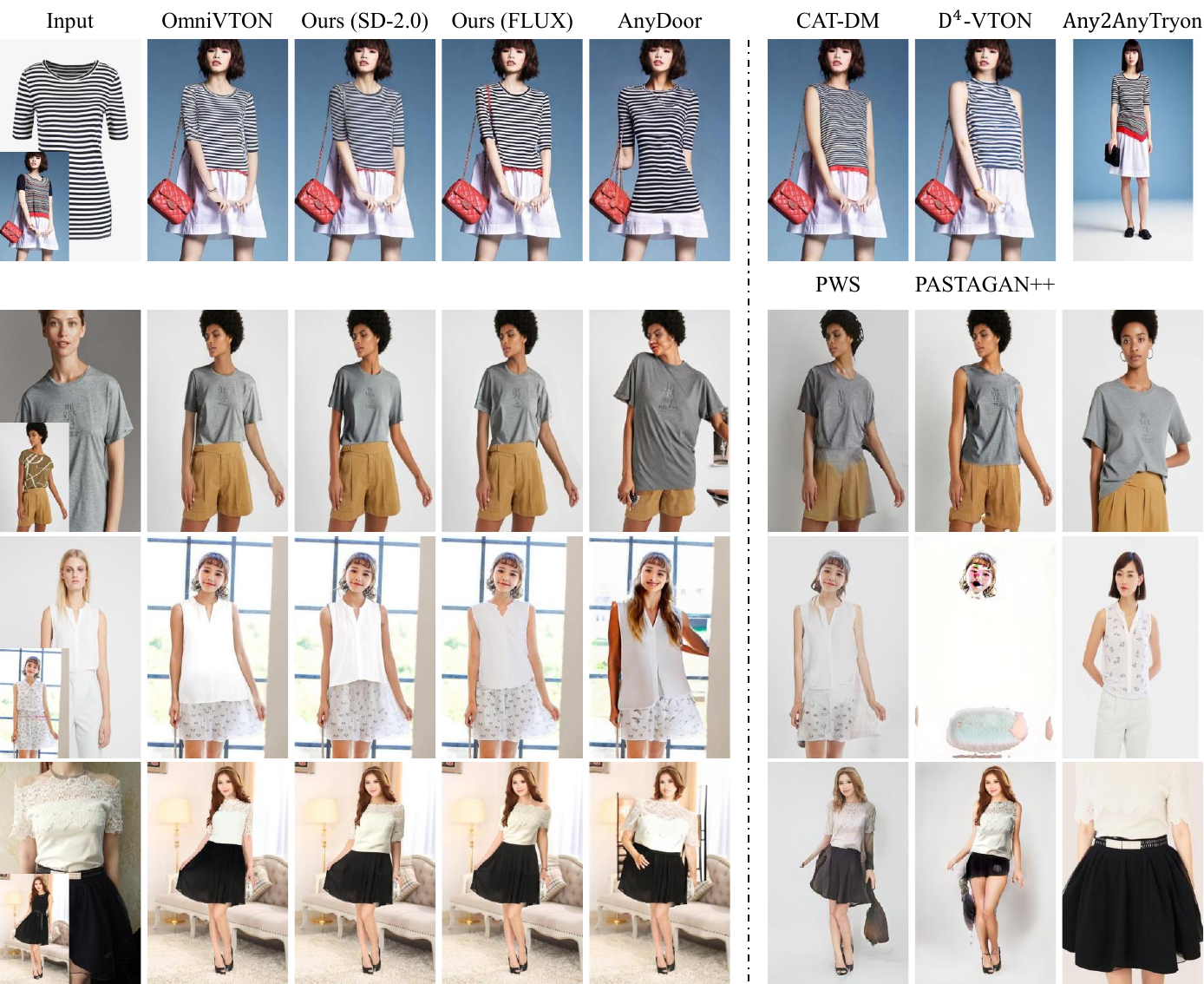}
	\captionof{figure}{Qualitative results on the StreetTryOn benchmark across different scenarios. From top to bottom: Shop-to-Street, Model-to-Model, Model-to-Street, and Street-to-Street.}
	\label{fig:street}
\end{figure}

Fig.~\ref{fig:street} presents qualitative comparisons on the StreetTryOn benchmark. General-purpose methods such as Any2AnyTryOn and AnyDoor demonstrate adaptability to different scenarios, but often fail to maintain garment texture and human pose consistency between the given person image and the try-on result. Methods trained on limited data distributions, including PWS and PastaGAN++, show reduced robustness when applied to real-world images, while CAT-DM and D$^4$-VTON are restricted to in-shop garment inputs. In contrast, OmniVTON++ yields the most reliable qualitative performance across scenarios.

\subsection{Ablation Analysis}\label{sec:ablation}
We conduct a component-wise macro ablation to assess the contribution of individual modules in OmniVTON++ using SD-2.0 and FLUX, with results evaluated on VITON-HD in Tab.~\ref{tab:ablation}. We start from a base model without any proposed modules (Base for SD-2.0 and Base* for FLUX), and incrementally integrate the proposed components to form variants (A)–(C) for SD-2.0 and (D)–(G) for FLUX. Specifically, variants (A)/(D) incorporate SGM, (B)/(E) incorporate PPG, (C)/(F) combine SGM with CBS, and variant (G) combines SGM with CBS-DiT. We next analyze the effectiveness of each module.

\begin{figure*}
	\centering
	\includegraphics[width=\linewidth]{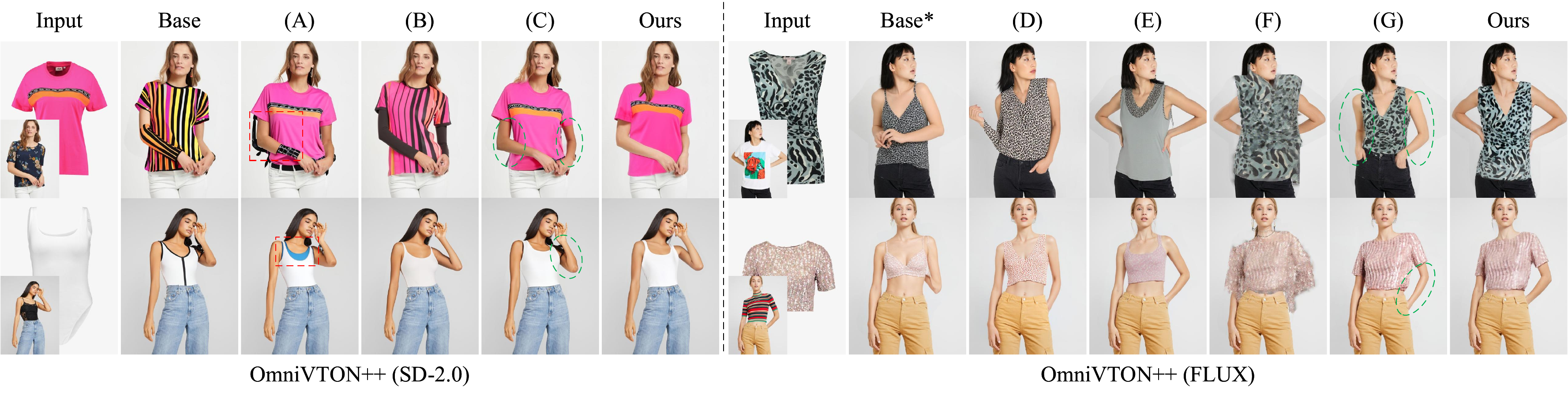}
	\captionof{figure}{Qualitative ablation results of component variants under SD-2.0 (left) and FLUX (right). Compared with individual or partial configurations, the complete model yields more accurate garment geometry and more coherent body structure.} 
	\label{fig:ablation}
\end{figure*}

\noindent\textbf{Effectiveness of SGM.}
As shown in Tab.~\ref{tab:ablation}, incorporating SGM results in substantial performance improvements over Base under the U-Net backbone, as reflected by gains across all evaluation metrics in variant (A). The qualitative results in Fig.~\ref{fig:ablation} further indicate that SGM provides body-aligned geometry and accurate texture patterns for the garment, while minor inpainting artifacts may still appear near part transition regions (red boxes).

When applied under DiT (variant (D)), SGM continues to improve performance relative to Base*, although the relative gain is smaller than that observed for variant (A). We attribute this difference to the architectural inductive biases that govern feature aggregation. Under the U-Net setting, spatial organization is predominantly shaped by convolutional operations, which favor local continuity and help preserve the structured cues encoded in the coarse morphed garment $I_w$. In contrast, DiT relies primarily on global self-attention, where garment features are integrated together with text tokens within a shared attention space. When $I_w$ contains boundary discontinuities, its representation deviates from the natural image manifold and is less reliably propagated under attention-based aggregation, making garment features weaker as structural references under joint garment-text conditioning, as also observed in the qualitative results.

\begin{table}[t]\centering
\def\arraystretch{1.2}
\small
\scriptsize
\tabcolsep 1.5pt
\caption{Macro ablation of Structured Garment Morphing (SGM), Continuous Boundary Stitching (CBS), and Spectral Pose Injection (SPI) on VITON-HD under the SD-2.0 and FLUX backbones. Best results are marked in \textbf{bold} per block.}
\resizebox{\linewidth}{!}{
\begin{tabular}{l c c c c c c c c c c c c}
  \toprule
  Method && Backbone && SGM & PPG & CBS & CBS-DiT && FID$_u$$\downarrow$ & FID$_p$$\downarrow$ & SSIM$_p$$\uparrow$ & LPIPS$_p$$\downarrow$ \\
  \cmidrule{1-1} \cmidrule{3-3} \cmidrule{5-8} \cmidrule{10-13}

  Base && SD-2.0 && & & & && 18.445 & 16.878 & 0.773 & 0.222 \\
  \cmidrule{1-1} \cmidrule{3-3} \cmidrule{5-8} \cmidrule{10-13}
  (A) && SD-2.0&& \ding{51} & & & && 12.946 & 10.949 & 0.803 & 0.173 \\
  (B) && SD-2.0&& & \ding{51} & & && 12.319 & 10.604 & 0.827 & 0.167 \\
  (C) && SD-2.0 && \ding{51} & & \ding{51} & && 9.331 & 7.484 & 0.821 & 0.154 \\
  \cmidrule{1-1} \cmidrule{3-3} \cmidrule{5-8} \cmidrule{10-13}
  \textbf{Ours} &&SD-2.0 && \ding{51} & \ding{51} & \ding{51} & && \textbf{9.189} & \textbf{6.990} & \textbf{0.843} & \textbf{0.130} \\
  \midrule  
  Base* &&FLUX && & & & && 13.900 & 11.564 & 0.827 & 0.167 \\
  \cmidrule{1-1} \cmidrule{3-3} \cmidrule{5-8} \cmidrule{10-13}
  (D) &&FLUX && \ding{51} & & & && 12.460 & 10.744 & 0.828 & 0.165 \\
  (E) &&FLUX && & \ding{51} & & && 10.410 & 8.946 & 0.844 & 0.145 \\
  (F) &&FLUX && \ding{51} & & \ding{51} & && 41.002 & 39.505 & 0.800 & 0.229 \\
  (G) &&FLUX && \ding{51} & & & \ding{51} && 10.396 & 7.836 & 0.841 & 0.136 \\
  \cmidrule{1-1} \cmidrule{3-3} \cmidrule{5-8} \cmidrule{10-13}
  \textbf{Ours} && FLUX&& \ding{51} & \ding{51} & & \ding{51} && 
  \textbf{9.286} & \textbf{6.618} & \textbf{0.849} & \textbf{0.121} \\
  \bottomrule
\end{tabular}
}
\label{tab:ablation}
\end{table}

\begin{table}[t]
    \centering
    \def\arraystretch{1.2}
    \small
    \scriptsize
    \tabcolsep 1.5pt
    \caption{Ablation study on alternative designs for SGM. Subscripts $u$ and $p$ denote the unpaired and paired settings, respectively. The best results are marked in \textbf{bold}.}
    \label{tab:ablation_sgm}
    \resizebox{\linewidth}{!}{
    \begin{tabular}{lc cc cccc}
        \toprule
        Variant && Backbone && FID$_u$$\downarrow$ & FID$_p$$\downarrow$ & SSIM$_p$$\uparrow$ & LPIPS$_p$$\downarrow$ \\
        \cmidrule{1-1} \cmidrule{3-3} \cmidrule{5-8}   
        Outpainting-based~\cite{omnivton} && SD-2.0 && 9.470 & 7.429 & 0.842 & 0.135 \\
        DensePose Flow-based && SD-2.0 && 12.043 & 10.540 & 0.840 & 0.145 \\       
        \cmidrule{1-1} \cmidrule{3-3} \cmidrule{5-8}
        \textbf{SGM (OmniVTON++)} && SD-2.0 && \textbf{9.189} & \textbf{6.990} & \textbf{0.843} & \textbf{0.130} \\
        \bottomrule
    \end{tabular}}
\end{table}

\begin{figure}
	\centering
	\includegraphics[width=\linewidth]{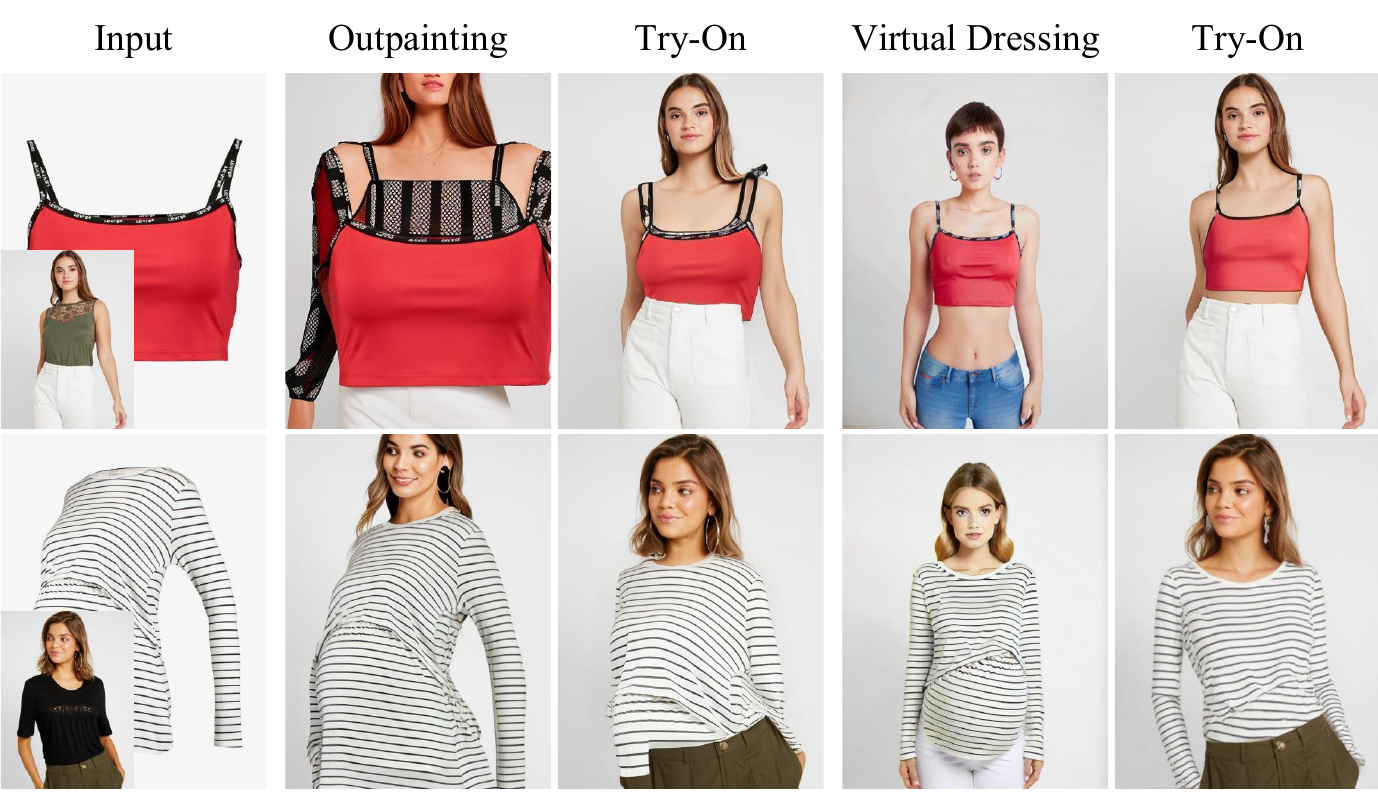}
	\captionof{figure}{Qualitative comparisons of pseudo-person image generation and downstream try-on results using attention-modulated outpainting (OmniVTON) and virtual dressing (OmniVTON++).}
	\label{fig:sgm_compare}
\end{figure}

To validate the necessity of individual design choices in SGM, we further perform a finer-grained ablation study on VITON-HD, where the proposed SGM is compared against two alternatives, with results summarized in Tab.~\ref{tab:ablation_sgm}. We first examine pseudo-person image generation by using the attention-modulated outpainting from OmniVTON instead of our virtual dressing module. Quantitative comparisons show that this alternative performs worse than ours across all metrics, implying that the resulting pseudo-person images introduce geometric ambiguities that affect garment morphing. Fig.~\ref{fig:sgm_compare} presents two representative qualitative examples, illustrating that the outpainting-based strategy can produce pseudo-person images in which garment-driven geometric patterns are misinterpreted as body structure. By contrast, virtual dressing relies on a learned prior with explicit human structural conditioning, and therefore avoids introducing such ambiguities.

The other alternative replaces the correspondence and transformation step in SGM with DensePose flow, which uses dense correspondences defined in the universal IUV space to warp garment pixels from the pseudo-person image onto the input person. As reported in Tab.~\ref{tab:ablation_sgm}, this strategy yields inferior performance for virtual try-on. Without task-specific retraining, the predicted UV fields are not optimized for garment-level alignment, which limits the preservation of garment identity in the synthesized results.

\noindent\textbf{Effectiveness of PPG.} Variants (B) and (E) in Tab.~\ref{tab:ablation} report the quantitative results of incorporating PPG into the base models, which leads to improvements across all evaluation metrics, with the relative gain on SSIM$_p$ being the most pronounced among different settings, indicating stronger preservation of global structural fidelity. This behavior arises from the use of the proxy latent, which conveys pose information and enables persistent control during generation, thereby preserving human body structure while avoiding interference from the original garment appearance. Qualitative comparisons in Fig.~\ref{fig:ablation} further show that pose misalignments, such as local limb inconsistencies (green circles), are reduced across both diffusion models, yielding results that better follow the given human pose.
\begin{table}[t]
    \centering
    \def\arraystretch{1.2}
    \small
    \scriptsize
    \tabcolsep 1.5pt
    \caption{Ablation study on pose guidance designs. Subscripts $u$ and $p$ denote the unpaired and paired settings, respectively. The best and second-best results are marked in \textbf{bold} and \underline{underlined}.}
    \label{tab:ablation_ppg}
    \resizebox{\linewidth}{!}{
    \begin{tabular}{lc cc cccc}
        \toprule
        Variant && Backbone && FID$_u$$\downarrow$ & FID$_p$$\downarrow$ & SSIM$_p$$\uparrow$ & LPIPS$_p$$\downarrow$ \\
        \cmidrule{1-1} \cmidrule{3-3} \cmidrule{5-8}        
        ControlNet~\cite{controlnet} && SD-2.0 && 10.910 & 8.778 & 0.818 & 0.159 \\ 
        SPI~\cite{omnivton} && SD-2.0 && \underline{9.365} & \textbf{6.964} & 0.832 & 0.141 \\
        Full-Latent &&SD-2.0 && 9.462 & 7.244 & \textbf{0.847} & \textbf{0.128} \\
        Low-Frequency Latent&& SD-2.0&& 9.514 & 7.431 & 0.835 & 0.139 \\       
        \cmidrule{1-1} \cmidrule{3-3} \cmidrule{5-8}
        \textbf{PPG (OmniVTON++)} && SD-2.0 && \textbf{9.189} & \underline{6.990} & \underline{0.843} & \underline{0.130} \\
        \bottomrule
    \end{tabular}}
\end{table}

We additionally conduct an ablation study on alternative pose guidance designs, as summarized in Tab.~\ref{tab:ablation_ppg}. These variants explore different strategies for pose conditioning during generation. ControlNet~\cite{controlnet} represents a widely used approach that injects pose constraints through an external control branch. However, prior studies~\cite{emma} have shown that jointly conditioning on multiple modalities (\eg, pose cues and text prompts) can lead to biased conditioning behavior, which adversely affects generation quality. In contrast, SPI, as proposed in OmniVTON~\cite{omnivton}, applies pose guidance only at initialization. Although both approaches can improve pose adherence, their effectiveness remains limited in terms of structural fidelity and perceptual quality.
 
We then focus on two variants that differ in the components of the intermediate prediction used to compute the residual with respect to the proxy latent. One variant, denoted Full-Latent, applies guidance to the entire intermediate prediction, while the other, denoted Low-Frequency Latent, restricts guidance to its low-frequency component. Both variants outperform ControlNet and SPI on pose-oriented metrics such as SSIM$_p$ and LPIPS$_p$. However, their FID scores are inferior to SPI, indicating reduced visual realism. For Full-Latent guidance, the proxy latent encodes minimal, non-discriminative garment cues; guiding the entire intermediate prediction therefore over-regularizes generation and degrades garment fidelity. Regarding Low-Frequency Latent guidance, it adopts a fixed truncation frequency following SPI, yet this choice can be suboptimal, as its effect may be sensitive across diffusion steps, making the selection of a single truncation frequency non-trivial. PPG applies pose guidance to the principal components of the intermediate prediction, enabling targeted pose control while leaving sufficient capacity for garment appearance modeling. As a result, it achieves a favorable trade-off between visual quality and structural accuracy among the compared variants. 
 
\begin{figure}
	\centering
	\includegraphics[width=\linewidth]{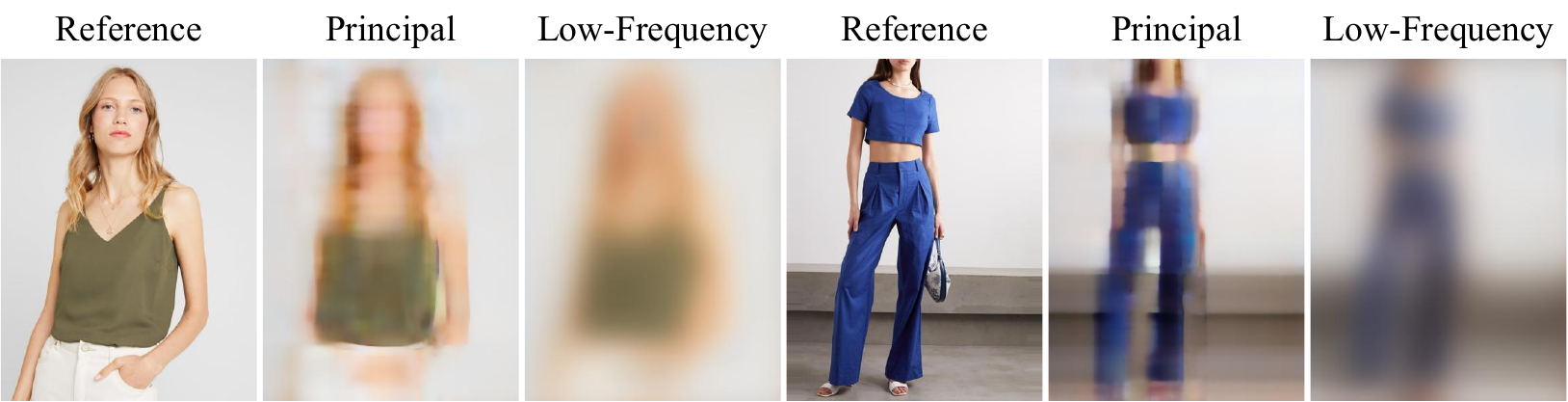}
	\captionof{figure}{Visualization of principal and low-frequency latent components. The principal components reveal a more compact encoding of pose-related structure.}
	\label{fig:low_freq}
\end{figure}

Fig.~\ref{fig:low_freq} compares the information captured by the principal components and the low-frequency components of the latent representation for a reference image. The principal components exhibit a more compact organization of pose-related structure, whereas the low-frequency components show a more diffuse distribution. This difference supports the use of principal components as a more reliable carrier of pose information within PPG.

\noindent\textbf{Effectiveness of CBS and CBS-DiT.}
As CBS operates on the garment-infused person image, where the garment content is provided by SGM, we evaluate their joint configuration in the macro ablation, corresponding to variants (C) and (F) in Tab.~\ref{tab:ablation}. Under the U-Net backbone, variant (C) yields clear quantitative improvements, which are further corroborated by the qualitative comparisons in Fig.~\ref{fig:ablation}. In particular, CBS improves local appearance coherence by suppressing stitching artifacts and restoring fine-grained texture continuity. When applied under the DiT backbone, however, directly using CBS (variant (F)) results in clear failure cases, as positional ambiguity in the shared positional space leads to misaligned correspondences that produce conflicting garment structures and severe visual artifacts, as illustrated in Fig.~\ref{fig:ablation}. In contrast, the adapted CBS-DiT variant (G) re-establishes a continuous positional space through PIR, allowing CBS to be applied over spatially coherent context. As a result, the artifacts observed in variant (F) are no longer present, and the visual quality is recovered.

\begin{table}[t]
    \centering
    \def\arraystretch{1.2}
    \small
    \scriptsize
    \tabcolsep 1.5pt
    \caption{Ablation study on alternative designs for CBS and CBS-DiT. Subscripts $u$ and $p$ denote the unpaired and paired settings, respectively. Best results are marked in \textbf{bold} per block.}
    \label{tab:ablation_cbs}
    \resizebox{\linewidth}{!}{
    \begin{tabular}{lc cc cccc}
        \toprule
        Variant && Backbone && FID$_u$$\downarrow$ & FID$_p$$\downarrow$ & SSIM$_p$$\uparrow$ & LPIPS$_p$$\downarrow$ \\
        \cmidrule{1-1} \cmidrule{3-3} \cmidrule{5-8}
        \rowcolor[HTML]{F2F2F2}
        \multicolumn{8}{l}{\textit{Continuous Boundary Stitching (CBS)}} \\
        Attention Reweighting~\cite{zstar} && SD-2.0 && 13.143 & 11.048 & 0.823 & 0.156 \\       
        Delta Denoising Score~\cite{freecompose} && SD-2.0 && 12.125 & 10.149 & 0.829 & 0.154 \\
        Asymmetric Interaction && SD-2.0 && 9.324 & 7.188 & 0.841 & 0.131 \\ 
        \cmidrule{1-1} \cmidrule{3-3} \cmidrule{5-8}
        \textbf{CBS (OmniVTON++)} && SD-2.0 && \textbf{9.189} & \textbf{6.990} & \textbf{0.843} & \textbf{0.130} \\
        \midrule       
        \cmidrule{1-1} \cmidrule{3-3} \cmidrule{5-8}
        \rowcolor[HTML]{F2F2F2}
        \multicolumn{8}{l}{\textit{Continuous Boundary Stitching for DiT (CBS-DiT)}} \\
        Value Suppression && FLUX && 10.890 & 8.536 & \textbf{0.855} & 0.128 \\
        Diptych Generation~\cite{diptych} && FLUX && 10.313 & 7.315 & 0.839 & 0.126 \\
        \cmidrule{1-1} \cmidrule{3-3} \cmidrule{5-8}
        \textbf{CBS-DiT (OmniVTON++)} && FLUX && \textbf{9.286} & \textbf{6.618} & 0.849 & \textbf{0.121} \\
        \bottomrule
    \end{tabular}}
\end{table}

We further study alternative boundary refinement strategies for CBS and CBS-DiT, with quantitative results reported in Tab.~\ref{tab:ablation_cbs}. For CBS, Attention Reweighting~\cite{zstar} incorporates garment information in this setting through garment-conditioned cross-attention in the person stream, alongside self-attention over person features. While garment cues enter the person-side attention computation, their features are not shared within a unified attention pool. This limits the model’s ability to reconcile garment and person evidence at boundary regions, leading to less accurate refinement.

Delta Denoising Score~\cite{freecompose} introduces garment cues in this setting by leveraging the discrepancy between the predicted noises of the person and garment streams at each denoising step to modulate the person-stream prediction. This modulation operates at the noise level rather than in a shared feature space with attention-based aggregation, leaving garment and person representations under separate distributions. It is therefore not structured to form garment–body associations required to resolve spatially mixed cues within the inpainting region, which limits garment appearance formation.

Asymmetric Interaction follows the CBS formulation but removes feedback from the person stream to the garment stream, restricting information flow to a single direction. In this case, garment features cannot be adjusted based on the current person representation and are reused as static references. Boundary stitching therefore becomes less adaptive to appearance variations introduced by the human body, leading to inferior performance compared with CBS.

For CBS-DiT, Value Suppression adopts the same feature aggregation strategy used in CBS. While this operation largely blocks irrelevant responses, it can disrupt positional continuity under DiT, leading to degraded performance. Diptych Generation~\cite{diptych}, when used as an alternative to CBS-DiT, concatenates person and garment tokens into a single sequence at the outset and processes them jointly. As a result, both inputs pass through a shared normalization and feed-forward path, which alters feature distributions prior to attention and undermines attention computation. By contrast, CBS-DiT keeps the two streams separate and introduces interaction only at the attention level, where similarity-based association is preserved, with attenuation over irrelevant responses. This design therefore leads to improved garment–person integration across both unpaired and paired settings.

\subsection{Multi-Garment Virtual Try-On}
Beyond the standard single-garment setting, we evaluate OmniVTON++ in more involved application scenarios. In particular, we consider multi-garment virtual try-on, where multiple garments are replaced for the same person. This setting is supported by OmniVTON++ without changes to its formulation, requiring only a reorganization of the inputs.

\begin{figure}
	\centering
	\includegraphics[width=\linewidth]{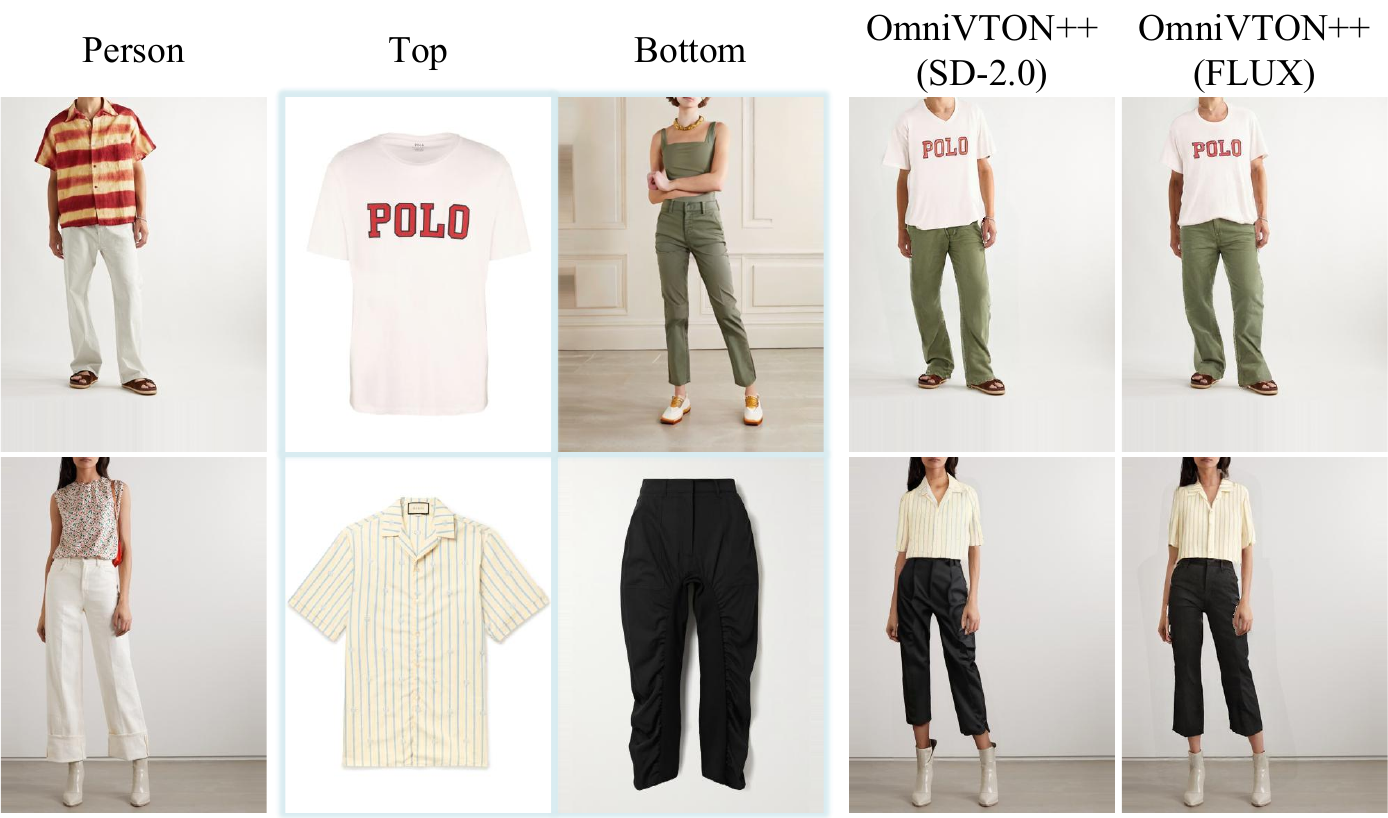}
	\captionof{figure}{Qualitative results of multi-garment virtual try-on with OmniVTON++.}
	\label{fig:multi_garment}
\end{figure}

Given multiple target-garment-wearing images, we concatenate them spatially, after which the composite is resized to match the resolution of the person image, forming the garment-stream input. In parallel, each target-garment-wearing image is morphed independently, and the resulting garment representations are injected into the person image to produce the garment-infused image. Qualitative results are shown in Fig.~\ref{fig:multi_garment}. This setting brings virtual try-on to the outfit level, enabling holistic preview to styling and acquisition decisions.

\subsection{Multi-Human Virtual Try-On}
We next consider a multi-human virtual try-on scenario, which remains unexplored within existing virtual try-on frameworks. In this setting, multiple people appear in the same person image, and identical garments may be assigned to different individuals, or distinct garments may be specified per person. Garment inputs follow the same organization as in our previous settings: identical garments are handled as in the standard single-garment case, while multiple garments use the multi-garment input formulation. As garment transformation and structural alignment in OmniVTON++ are performed on a per-instance basis, our formulation accommodates multiple subjects without training adaptation. The resulting image contains multiple garment-infused subjects, with inpainted regions subsequently completed and refined.

\begin{figure}
	\centering
	\includegraphics[width=\linewidth]{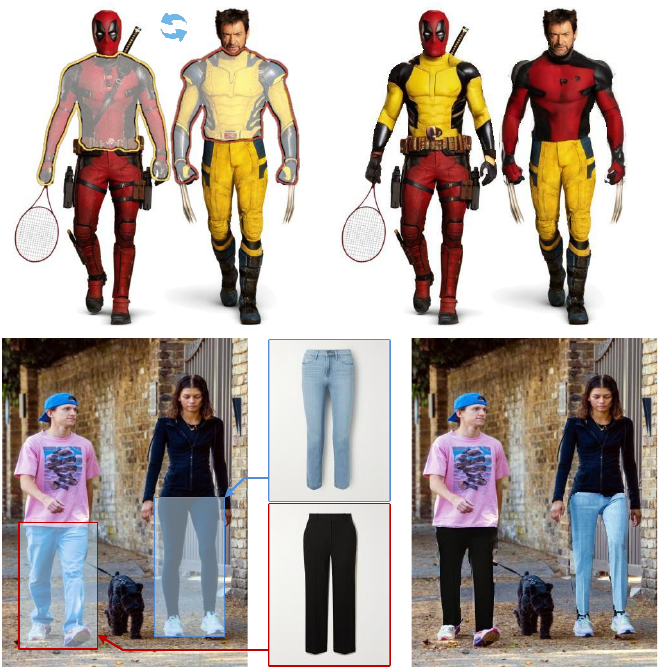}
	\captionof{figure}{Multi-human virtual try-on results. Top row shows Model-to-Model, while bottom row depicts Shop-to-Street.}
	\label{fig:multi_user}
\end{figure}

Qualitative results using SD-2.0 are shown in Fig.~\ref{fig:multi_user}, demonstrating the applicability of OmniVTON++ on multi-human configurations across heterogeneous input regimes. This extension broadens the scope of virtual try-on tasks, enabling group-centric fashion applications such as coordinated outfits and uniform design.

\subsection{Anime Character Virtual Try-On}

\begin{figure}
	\centering
	\includegraphics[width=\linewidth]{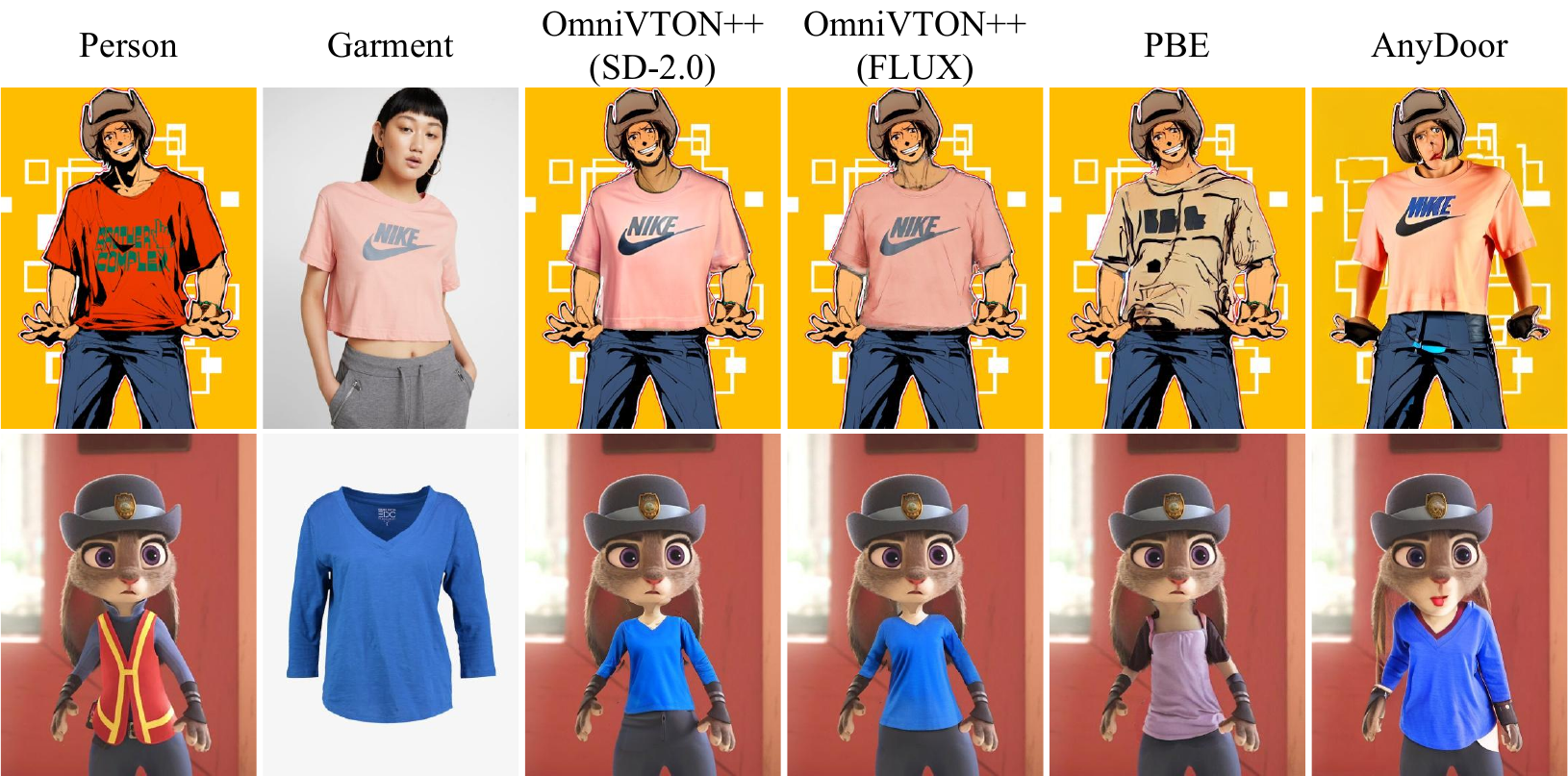}
	\captionof{figure}{Anime character virtual try-on results. Comparison of OmniVTON++ with different backbones and representative baselines on stylized character inputs.}
	\label{fig:anime_style}
\end{figure}

We evaluate OmniVTON++ on anime character virtual try-on, where the person images depict illustrated characters rather than real humans. Compared with other scenarios, the primary difference lies in visual statistics and artistic rendering style. Our method requires no modification under this domain shift.

Qualitative comparisons in Fig.~\ref{fig:anime_style} include two exemplar-based editing baselines, PBE and AnyDoor. Both baselines tend to compromise either garment fidelity or character identity, whereas our framework preserves garment appearance and character consistency when adapting garments to anime characters. This behavior aligns with the formulation of OmniVTON++, which operates primarily on structural relationships rather than relying on photorealistic priors. Consequently, the generated results remain stable under stylized rendering. Such use cases arise in digital illustration and character design workflows, where virtual try-on streamlines outfit exploration and reduces manual effort during iterative development.

\section{Limitations}
\begin{figure}[t]
	\centering
	\includegraphics[width=\linewidth]{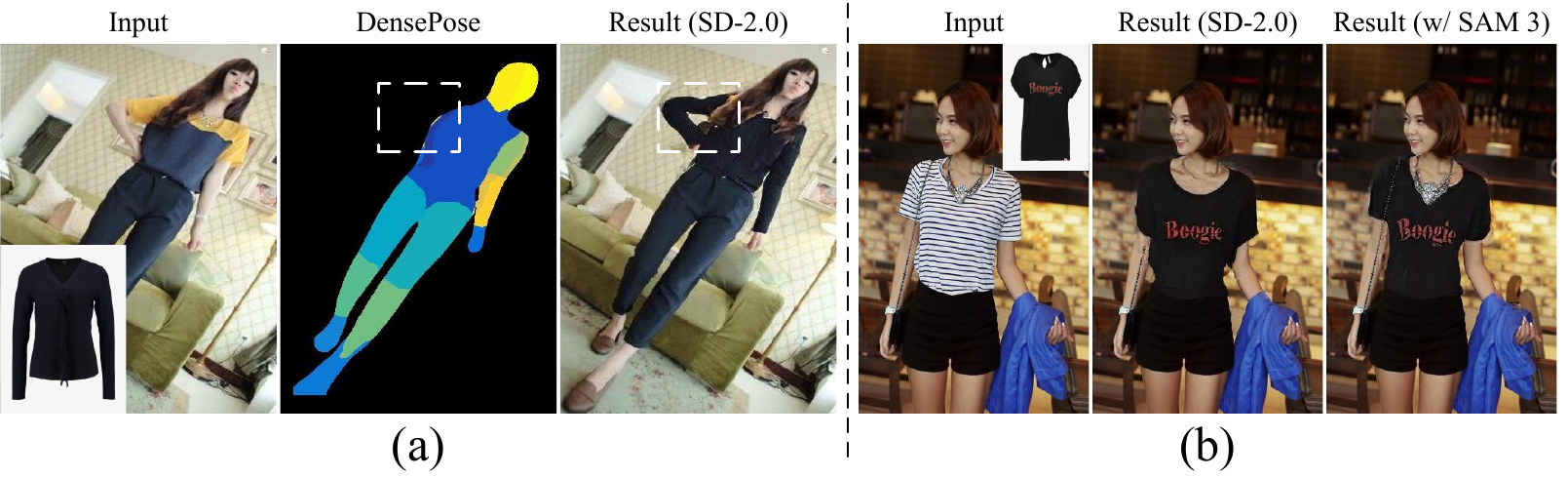}
	\captionof{figure}{Failure cases of OmniVTON++. (a) Imprecise human analysis signals. (b) Incorrect accessory masking.}
	\label{fig:limitation}
\end{figure}

Despite its universal formulation and demonstrated state-of-the-art performance, OmniVTON++ still has several limitations. First, the framework depends on the accuracy of external human analysis signals, including pose keypoint estimation, human parsing, and dense correspondence prediction. Inaccurate or missing keypoints can degrade the deformation quality of the SGM module and affect the construction of the agnostic mask, while imprecise human parsing or DensePose predictions weaken the structural guidance in PPG. As illustrated in Fig.~\ref{fig:limitation}(a), under large body rotations and cluttered backgrounds, DensePose fails to reliably localize semantic regions of the right arm, preventing our method from constructing an accurate proxy representation and thus producing distortions in the synthesized limb region.

Another limitation is the preservation of fine-grained accessories. Because the agnostic mask is derived from skeletal topology, accessories such as necklaces may be mistakenly included in the inpainting region and consequently removed, as shown in Fig.~\ref{fig:limitation}(b). This issue can be mitigated by incorporating general-purpose segmentation models. One such model is SAM 3~\cite{sam3}, in which text prompts are used to generate dedicated accessory masks to refine the masking process accordingly, enabling correct retention of accessory regions.

\section{Conclusion}
In this paper, we present OmniVTON++, a universal framework for virtual try-on that operates across a wide range of settings. The proposed approach separates garment geometry, appearance, and pose control, enabling stable behavior without relying on domain- or setting-specific training.

OmniVTON++ integrates Structured Garment Morphing to establish geometry-aligned garment representations, Continuous Boundary Stitching to regulate appearance transitions at garment boundaries, and Principal Pose Guidance to maintain pose consistency during generation. Together, these components support virtual try-on under diverse configurations within a single formulation.

Experimental results from standard benchmarks demonstrate effectiveness in cross-dataset and cross-garment-type settings, as well as across scenarios and diffusion backbones. Extended evaluations further validate our framework in multi-garment, multi-human, and anime character virtual try-on tasks within the same formulation. We view this work as a meaningful leap toward a more flexible and practically deployable virtual try-on system.

{
    \small
    \bibliographystyle{IEEEtran}
    \bibliography{main}
}

\end{document}